\def\paperID{0009}
\def\confName{IMW}
\def\confYear{2026}

\def\paperTitle{DeepShapeMatchingKit: Accelerated Functional Map Solver and \\Shape Matching Pipelines Revisited
}

\newif\ifreview 
\newif\ifarxiv \newcommand{\arxiv}{\arxivtrue}
\newif\ifcamera 
\newif\ifrebuttal 

\arxiv %

\pdfoutput=1
\documentclass[10pt,twocolumn,letterpaper]{article}
\ifreview \usepackage[review]{cvpr} \fi
\ifarxiv \usepackage[pagenumbers]{cvpr} \fi
\ifrebuttal \usepackage[rebuttal]{cvpr} \fi
\ifcamera \usepackage{cvpr} \fi

\usepackage{graphicx}	
\usepackage{amsmath}	
\usepackage{amssymb}	
\usepackage{booktabs}
\usepackage{times}
\usepackage{microtype}
\usepackage{epsfig}
\usepackage{caption}
\usepackage{float}
\usepackage{placeins}
\usepackage{color, colortbl}
\usepackage{stfloats}
\usepackage{enumitem}
\usepackage{tabularx}
\usepackage{xstring}
\usepackage{multirow}
\usepackage{xspace}
\usepackage{url}
\usepackage{subcaption}
\usepackage{xcolor}
\usepackage[hang,flushmargin]{footmisc}
\usepackage[ruled,vlined]{algorithm2e}

\ifcamera \usepackage[accsupp]{axessibility} \fi

\ifarxiv  \fi

\newcommand{\R}[1]{{%
    \textbf{%
        \ifstrequal{#1}{1}{\textcolor{red}{R#1}}{%
        \ifstrequal{#1}{2}{\textcolor{blue}{R#1}}{%
        \ifstrequal{#1}{3}{\textcolor{magenta}{R#1}}{%
        \ifstrequal{#1}{4}{\textcolor{teal}{R#1}}{%
                           \textcolor{cyan}{R#1}%
        }}}}%
    }%
}}

\usepackage{minted}
\usepackage{tcolorbox}
\tcbuselibrary{minted,skins,breakable}

\newtcblisting{codecard}[2][]{%
  listing engine=minted,
  listing only,            %
  minted language=#2,
  minted options={
    fontsize=\small,
    numbersep=6pt,
    tabsize=2,
    escapeinside=||
  },
  boxrule=0pt,
  colback=gray!7,
  arc=10pt,               %
  outer arc=10pt,
  left=10pt,right=10pt,top=10pt,bottom=10pt,
colframe=black!15,
boxrule=0.5pt,,
  #1
}

\usepackage{xr-hyper}

\makeatletter
\newcommand*{\addFileDependency}[1]{
  \typeout{(#1)}
  \@addtofilelist{#1}
  \IfFileExists{#1}{}{\typeout{No file #1.}}
}

\makeatother
\newcommand*{\myexternaldocument}[1]{
    \externaldocument{#1}
    \addFileDependency{#1.tex}
    \addFileDependency{#1.aux}
}

\definecolor{cvprblue}{rgb}{0.21,0.49,0.74}
\usepackage[pagebackref,breaklinks,colorlinks,allcolors=cvprblue]{hyperref}
\usepackage[capitalize]{cleveref}
\crefname{section}{Sec.}{Secs.}
\crefname{table}{Table}{Tables}
\crefname{figure}{Fig.}{Figs.}

\ifarxiv \crefname{appendix}{App.}{Apps.}
\else \crefname{appendix}{Suppl.}{Suppls.} \fi

\definecolor{cvprblue}{rgb}{0.21,0.49,0.74}
\definecolor{cPLOT_gc_ppsm}{RGB}{127, 201, 127}
\definecolor{cPLOT_sm_comb}{RGB}{253, 192, 10}
\definecolor{cPLOT_dpfm}{RGB}{240, 2, 127}
\definecolor{cPLOT_ours}{RGB}{20, 120, 120}
\definecolor{cPLOT5}{RGB}{39, 174, 239}
\definecolor{cPLOT6}{RGB}{179,0,0}
\definecolor{cPLOT7}{RGB}{199,21,133}
\definecolor{cPLOT8}{RGB}{30,144,255}

\usepackage[dvipsnames]{xcolor}

\usepackage[capitalize]{cleveref}
\crefname{section}{Sec.}{Secs.}
\crefname{table}{Table}{Tables}
\crefname{figure}{Fig.}{Figs.}

\usepackage{graphicx}	
\usepackage{amsmath}	
\usepackage{amssymb}	
\usepackage{booktabs}
\usepackage{times}
\usepackage{microtype}
\usepackage{epsfig}
\usepackage[table,xcdraw,dvipsnames]{xcolor}
\usepackage{caption}
\usepackage{float}
\usepackage{placeins}
\usepackage{color, colortbl}
\usepackage{stfloats}
\usepackage{enumitem}
\usepackage{tabularx}
\usepackage{xstring}
\usepackage{multirow}
\usepackage{xspace}
\usepackage{url}
\usepackage{subcaption}
\usepackage{xcolor}
\usepackage[hang,flushmargin]{footmisc}

\usepackage{comment}
\usepackage{amsthm}

\usepackage{color}
\usepackage{soul}

\makeatletter
\renewcommand\footnoterule{%
  \kern-50pt%
  \hrule\@width.4\columnwidth
  \kern2.6pt%
}
\makeatother

\definecolor{yizheng}{rgb}{0.9,0.,0.5}

\definecolor{zorah}{rgb}{0.,0.6,0.3}

\newcommand{\YZ}[1]{\textcolor{yizheng}{{#1}}}

\definecolor{ForestGreen}{rgb}{0.13, 0.55, 0.13}
\usepackage{pifont}%
\newcommand{\cmark}{{\color{ForestGreen} \ding{51}}}%
\newcommand{\xmark}{{\color{red} \ding{55}}}%

\definecolor{TRColor}{RGB}{240, 240, 240}

\usepackage{cuted}

\AtEndPreamble{
    \crefname{theorem}{Thm.}{Thms.}
    \crefname{lemma}{Lemma}{Lemmas}
}
\usepackage[misc]{ifsym}

\definecolor{darkgreen}{HTML}{228C21}
\renewcommand{\top}{T}
\usepackage{makecell}%

\usepackage{amsfonts}      
\usepackage{nicefrac}       
\usepackage{microtype}     
\usepackage{xcolor}         
\usepackage{graphicx}
\usepackage{adjustbox}
\usepackage{comment}
\usepackage{amsmath}
\usepackage{rotating}
\usepackage{multirow}
\usepackage{array}
\usepackage{caption}
\usepackage{subcaption}
\usepackage{pgfplots}
\usepackage{tikzsymbols}
\pgfplotsset{compat=1.18} 
\usepackage[11pt]{moresize}
\usepackage[percentage]{overpic}
\usepackage{pifont}
\usepackage{bbm}

\usepackage{dsfont}
\usetikzlibrary{external}

\definecolor{forestgreen}{rgb}{0.13, 0.55, 0.13}
\definecolor{cPLOT1}{RGB}{214,113,176}
\definecolor{cPLOT3}{RGB}{80,150,80}
\definecolor{cPLOT4}{RGB}{165, 124, 27}
\definecolor{cPLOT2}{RGB}{68, 33, 175}
\definecolor{cPLOT5}{RGB}{39, 174, 239}
\definecolor{cPLOT6}{RGB}{179,0,0}
\definecolor{cPLOT7}{RGB}{199,21,133}
\definecolor{cPLOT8}{RGB}{30,144,255}

\newcolumntype{?}{!{\vrule width 1pt}}

\newcolumntype{?}{!{\vrule width 1pt}}

\unless\ifarxiv \myexternaldocument{_supplementary} \fi

\begin{document}
\title{\paperTitle}
\newcommand{\authorspace}{\hspace{0.7cm}}
\newcommand{\affiliationspace}{\hspace{0.68cm}}

\newcommand{\yz}[1]{\textcolor{red}{[YZ: #1]}}

\author{
Yizheng Xie$^{1,6}$ \quad
Lennart Bastian$^{1,2,5}$ \quad
Congyue Deng$^{3}$ \\
Thomas W. Mitchel$^{4}$ \quad
Maolin Gao$^{1,2}$ \quad
Daniel Cremers$^{1,2}$ \\ \\[-0.3em]
$^1$ Technical University of Munich \quad
$^2$ Munich Center for Machine Learning \\
$^3$ MIT \quad
$^4$ Adobe \quad
$^5$ Imperial College London \quad
$^6$ Simon Fraser University
}

\maketitle

\begin{abstract}

\noindent Deep functional maps, leveraging learned feature extractors and spectral correspondence solvers, are fundamental to non-rigid 3D shape matching. 
Based on an analysis of open-source implementations, we find that standard functional map implementations solve k independent linear systems serially, which is a computational bottleneck at higher spectral resolution. 
We thus propose a vectorized reformulation that solves all systems in a single kernel call, achieving up to a $33\times$ speedup while preserving the exact solution. 
Furthermore, we identify and document a previously unnoticed implementation divergence in the spatial gradient features of the mainstay DiffusionNet: two variants that parameterize distinct families of tangent-plane transformations, and present experiments analyzing their respective behaviors across diverse benchmarks. 
We additionally revisit overlap prediction evaluation for partial-to-partial matching and show that balanced accuracy provides a useful complementary metric under varying overlap ratios. 
To share these advancements with the wider community, we present an open-source codebase, \textit{DeepShapeMatchingKit}, that incorporates these improvements and standardizes training, evaluation, and data pipelines for common deep shape matching methods. The codebase is available at: \url{https://github.com/xieyizheng/DeepShapeMatchingKit}

\end{abstract}

\section{Introduction}
\label{sec:intro}

Shape matching is a long-standing problem in computer vision and computer graphics, with applications ranging from texture transfer~\cite{dinh2005texture} and shape interpolation~\cite{cao2024spectral} to animation~\cite{eisenberger_neuromorph_2021} and medical analysis~\cite{bastian_s3m_2023}. 
Among existing approaches, deep functional map methods, which combine the functional map framework~\cite{ovsjanikov2012functional} with learned feature extractors, have emerged as a foundational paradigm, achieving strong results across both complete and partial non-rigid 3D shape matching. 
Beyond shape matching, functional maps have also found applications in image matching~\cite{cheng2024zero} and latent space correspondence~\cite{fumero2024latent}, underscoring their generality and importance as a differentiable matching module within learned pipelines.

\begin{figure}[t!]
    \centering
    
    \newcommand{\specResLineWidth}{2pt}
\newcommand{\plotWidth}{1.0\columnwidth}
\newcommand{\plotHeight}{0.7\columnwidth}
\newcommand{\specResTitle}{Functional Map Solver}
\newcommand{\specTension}{0.3}
\newcommand{\specTextsize}{\small}

\pgfplotsset{
  label style = {font=\specTextsize},
  title style = {font=\normalsize},
  legend style={
    fill=gray!10,
    fill opacity=0.6,
    draw=gray!20,
    text opacity=1
  }
}

\begin{tikzpicture}[scale=0.85, transform shape]
  \begin{axis}[
    width=\plotWidth,
    height=\plotHeight,
    grid=major,
    title=\textsc{\textbf{\specResTitle}},
    legend style={
      at={(0.97,0.97)},
      anchor=north east,
      legend columns=1,
      font=\specTextsize,
      draw=none %
    },
    legend cell align={left},
    ylabel={{\specTextsize Runtime [ms]}},
    xlabel={{\specTextsize Number of eigenfunctions $k$}},
    xmin=20, xmax=300,
    xtick={100,200,300},
    ylabel near ticks,
    ymin=0, ymax=550,
    ytick={0,100,200,300,400,500},
  ]
\addplot[
  draw=none,
  fill=green!18,
    fill opacity=0,
  forget plot
] coordinates {
(20,12.17)
(25,15.07)
(30,18.48)
(35,22.72)
(40,25.21)
(45,28.69)
(50,31.97)
(55,35.82)
(60,40.18)
(65,43.63)
(70,48.44)
(75,53.05)
(80,57.80)
(85,63.27)
(90,67.57)
(95,72.38)
(100,78.46)
(105,83.85)
(110,89.92)
(115,96.39)
(120,101.81)
(125,108.37)
(130,116.38)
(135,122.57)
(140,129.47)
(145,138.17)
(150,163.79)
(155,152.70)
(160,156.01)
(165,168.68)
(170,176.22)
(175,184.58)
(180,193.45)
(185,202.07)
(190,210.54)
(195,225.67)
(200,233.57)
(205,243.60)
(210,253.21)
(215,263.23)
(220,273.46)
(225,283.69)
(230,293.98)
(235,303.98)
(240,313.79)
(245,328.35)
(250,337.33)
(255,348.77)
(260,425.29)
(265,440.78)
(270,456.07)
(275,473.56)
(280,487.10)
(285,503.61)
(290,520.57)
(295,535.24)
(300,551.43)
(300,17.46)
(290,16.53)
(280,14.66)
(270,13.47)
(260,12.29)
(250,10.94)
(240,9.54)
(230,9.28)
(220,8.39)
(210,7.86)
(200,7.05)
(190,6.03)
(180,5.57)
(170,5.23)
(160,4.45)
(150,4.33)
(140,3.85)
(130,3.89)
(120,2.96)
(110,2.80)
(100,2.59)
(90,2.37)
(80,2.65)
(70,2.79)
(60,1.90)
(50,1.82)
(40,1.98)
(30,1.11)
(20,1.98)
}\closedcycle;
  \addplot [color=red, smooth, tension=\specTension, line width=\specResLineWidth]
  table[row sep=crcr]{%
20 12.17\\
25 15.07\\
30 18.48\\
35 22.72\\
40 25.21\\
45 28.69\\
50 31.97\\
55 35.82\\
60 40.18\\
65 43.63\\
70 48.44\\
75 53.05\\
80 57.80\\
85 63.27\\
90 67.57\\
95 72.38\\
100 78.46\\
105 83.85\\
110 89.92\\
115 96.39\\
120 101.81\\
125 108.37\\
130 116.38\\
135 122.57\\
140 129.47\\
145 138.17\\
150 163.79\\
155 152.70\\
160 156.01\\
165 168.68\\
170 176.22\\
175 184.58\\
180 193.45\\
185 202.07\\
190 210.54\\
195 225.67\\
200 233.57\\
205 243.60\\
210 253.21\\
215 263.23\\
220 273.46\\
225 283.69\\
230 293.98\\
235 303.98\\
240 313.79\\
245 328.35\\
250 337.33\\
255 348.77\\
260 425.29\\
265 440.78\\
270 456.07\\
275 473.56\\
280 487.10\\
285 503.61\\
290 520.57\\
295 535.24\\
300 551.43\\
  };

  \addplot [color=ForestGreen, smooth, tension=\specTension, line width=\specResLineWidth]
  table[row sep=crcr]{%
20 1.98\\
30 1.11\\
40 1.98\\
50 1.82\\
60 1.90\\
70 2.79\\
80 2.65\\
90 2.37\\
100 2.59\\
110 2.80\\
120 2.96\\
130 3.89\\
140 3.85\\
150 4.33\\
160 4.45\\
170 5.23\\
180 5.57\\
190 6.03\\
200 7.05\\
210 7.86\\
220 8.39\\
230 9.28\\
240 9.54\\
250 10.94\\
260 12.29\\
270 13.47\\
280 14.66\\
290 16.53\\
300 17.46\\
  };

  \legend{Previous\cite{donati2020deep, attaiki2021dpfm, cao_unsupervised_2023, bastian2024hybrid,li_learning_2022,xie2025echomatch}, Ours}

  \end{axis}
\end{tikzpicture}
    \vspace{-2mm}
\caption{%
\textbf{Runtime vs.\ spectral resolution.}
Comparison between existing loop-based functional map solvers~\cite{donati2020deep, attaiki2021dpfm, cao_unsupervised_2023, bastian2024hybrid,li_learning_2022,xie2025echomatch}
and our batched implementation.
Our batched solver achieves a $33\times$ speedup while preserving exact solutions.%
}
\vspace{-2mm}
\label{fig:spectral_runtime}
\vspace{-4.0mm}
\end{figure}
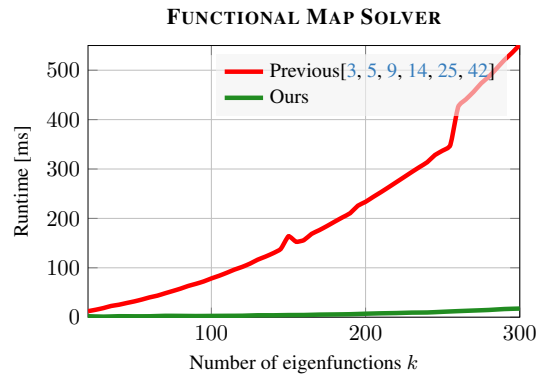

A central computational step in these pipelines is solving a 
regularized least-squares system to recover the functional map 
from learned spectral descriptors. The standard approach, first 
introduced in GeomFmaps~\cite{donati2020deep}, decouples 
the system row-wise and solves $k$ independent $k \times k$ 
linear systems in a loop, where $k$ is the spectral resolution. 
While conceptually simple, this loop becomes a significant 
bottleneck as $k$ increases. We observe that these independent 
systems can be reformulated as a single batched tensor solve, 
producing the exact same solution with up to a $33\times$ 
runtime speedup, see Fig.~\ref{fig:spectral_runtime}.

\begin{figure*}[t]
    \centering
    \vspace{-2mm}
    \includegraphics[width=\textwidth]{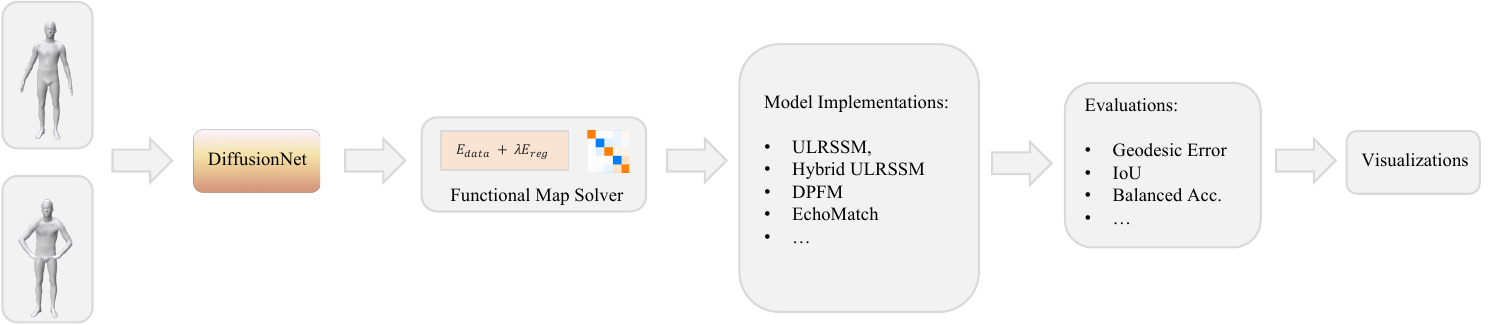}
\vspace{-5.0mm}
\caption{%
\textbf{Overview of common deep functional map pipelines.}
A shared pipeline underlies most deep shape matching methods:
a feature backbone, e.g., DiffusionNet, extracts per-vertex descriptors,
a functional map solver,
and method-specific components produce the final matching.
We revisit three parts of this pipeline:
a batched functional map solver,
an analysis of DiffusionNet variants,
and balanced accuracy as a complementary overlap
prediction metric.%
}

    \label{fig:dsmk_teaser}
\vspace{-4.0mm}
\vspace{-1mm}
\end{figure*}

In the process of unifying multiple deep shape matching methods 
into a common framework, see Table~\ref{tab:framework_comparison} and Fig~\ref{fig:dsmk_teaser}, we further identify a previously 
undocumented implementation divergence in DiffusionNet~\cite{sharp2022diffusionnet}, 
the de facto feature backbone for deep functional map pipelines. 
Two variants, stemming from a subtle difference in the 
computation of spatial gradient features, parameterize distinct 
families of tangent-plane transformations, and have been used 
in parallel across the literature~\cite{sharp2022diffusionnet, attaiki2021dpfm, xie2025echomatch,li_learning_2022,donati2022deep,zeng2025coe,attaiki2023understanding,cao_unsupervised_2023, bastian2024hybrid,vigano2025geomfum,cao2024spectral,cao2022unsupervised,cao_self-supervised_2023,ehm2024partial,roetzer2024spidermatch} 
without explicit documentation of their differences.

Finally, we revisit overlap prediction evaluation for 
partial-to-partial shape matching. Standard metrics such as IoU 
are sensitive to the overlap ratio, making comparisons across 
shape pairs with varying overlap difficult. We show that 
balanced accuracy, widely used in imbalanced classification but 
not previously adopted in this setting, provides a useful 
complementary measure.

We summarize our contributions as follows:
\begin{itemize}
    \item A batched functional map solver achieving up to $33\times$ speedup while preserving the exact solution, applicable to any deep functional map pipeline.
    \item Documentation and experimental analysis of two silently diverged DiffusionNet spatial gradient variants, revealing complementary behaviors across deformation settings.
    \item Balanced accuracy as a complementary overlap prediction metric for partial-to-partial shape matching, better isolating predictive quality from class imbalance.
\end{itemize}

We release our implementations as part of \textit{DeepShapeMatchingKit}, an open-source codebase providing unified training, evaluation, and data pipelines for common deep shape matching methods.

\section{Related Work}
\label{sec:related}

\begin{table*}[tbh!]
\vspace{-4mm}
\centering
\small
\begin{tabular}{@{}lcccccccc@{}}
\toprule
Frameworks & Fast Solver & Diff. (A) & Diff. (B) & \  \  \ F2F \  \  \  & \  \  \ P2F \  \  \  & \  \  \ P2P \  \  \  & \  \  IoU  \  \ & Bal.\ Acc \\ \midrule
PyFM~\cite{pyfmaps2025}               & \xmark & \xmark & \xmark & \cmark & \cmark & \xmark & \xmark & \xmark \\
GeomFUM~\cite{vigano2025geomfum}       & \xmark & \xmark & \cmark & \cmark & \cmark & \xmark & \xmark & \xmark \\
ULRSSM~\cite{cao_unsupervised_2023}    & \xmark & \xmark & \cmark & \cmark & \cmark & \xmark & \xmark & \xmark \\
Hybrid FMaps~\cite{bastian2024hybrid}  & \xmark & \xmark & \cmark & \cmark & \xmark & \xmark & \xmark & \xmark \\
AttentiveFmaps~\cite{li_learning_2022} & \xmark & \cmark & \xmark & \cmark & \xmark & \xmark & \xmark & \xmark \\
DPFM~\cite{attaiki2021dpfm}            & \xmark & \cmark & \xmark & \cmark & \cmark & \cmark & \cmark & \xmark \\
EchoMatch~\cite{xie2025echomatch}      & \xmark & \cmark & \xmark & \xmark & \xmark & \cmark & \cmark & \xmark \\
Ours                                   & \cmark & \cmark & \cmark & \cmark & \cmark & \cmark & \cmark & \cmark \\ \bottomrule
\end{tabular}
\vspace{-1mm}
\caption{We present \textit{DeepShapeMatchingKit}, an open-source code base for accelerated FM computation, supporting both DiffusionNet variants, full-to-full (F2F), partial-to-full (P2F), and partial-to-partial (P2P) matching, 
and complementary overlap prediction metrics.}
\label{tab:framework_comparison}
\vspace{-4mm}
\vspace{-1mm}
\end{table*}

\noindent\textbf{Deep Shape Matching.}
Recent advances in deep shape matching build upon the functional map framework~\cite{ovsjanikov2012functional} by combining learned descriptors with deep functional map regularizations~\cite{litany_deep_2017,halimi_unsupervised_2019,sharp2022diffusionnet}. Unsupervised learning of deep functional maps was further explored in ULRSSM~\cite{cao_unsupervised_2023}, which demonstrated that functional map regularization coupled with soft point-wise map consistency can achieve competitive performance without ground-truth supervision. To better handle non-isometric deformations, Hybrid Functional Map~\cite{bastian2024hybrid} extends the functional map formulation by incorporating extrinsic elastic spectral bases. In the context of partial shape matching, DPFM~\cite{attaiki2021dpfm} introduces a supervised cross-attention mechanism for partial overlap prediction. More recently, EchoMatch~\cite{xie2025echomatch} further extends partial-to-partial matching by modeling soft round-trip correspondence reflection for overlap prediction.

\noindent\textbf{Functional Map Solver.}
Functional map based methods rely on solving regularized least squares systems with precomputed spectral operators, a step that can become computationally demanding at higher spectral resolutions~\cite{donati2020deep,attaiki2021dpfm,cao_unsupervised_2023}. Recent work Scalable Dense Maps~\cite{magnet2024memory} proposes a memory efficient alternative that replaces the explicit solver with differentiable refinement in the spatial and functional domains, trading a modest accuracy for reduced memory consumption and runtime. In a different direction, Neural Isometries~\cite{mitchel2024neural}, while not directly applied to shape matching, explores learning orthogonal transformations via SVD based functional map constraints in a latent space, enabling fast approximate solvers on the orthogonal group, but deviating from the standard regularized functional map formulation. Inspired by but in contrast to these approaches, we focus on accelerating the original regularized functional map solution itself. %

\noindent\textbf{DiffusionNet Implementation.}
DiffusionNet~\cite{sharp2022diffusionnet} introduces a discretization agnostic neural architecture for learning on surfaces and has become a de facto choice for many deep shape matching pipelines, where it is commonly used as a learned descriptor within functional map based frameworks. A key component of its design is the use of spatial gradient features, which enable the learning of directional filters and have been shown to be effective for shape matching. Over time, DiffusionNet implementations have been adopted across a wide range of works~\cite{attaiki2021dpfm,cao2022unsupervised,donati2022deep,li_learning_2022,cao_unsupervised_2023,bastian2024hybrid,xie2025echomatch}. In particular, two variants, stemming from a difference in the computation of spatial gradient features from learned scaling and rotation, have been used in parallel in the literature. While both are highly effective, they exhibit subtle behavioral differences and are not checkpoint compatible. As part of integrating multiple deep shape matching pipelines into a unified framework, we feature both variants, provide documentation of their differences, and present experiments and analysis to demonstrate their respective behaviors.

\noindent\textbf{Overlap Prediction Metrics.}
Overlap prediction is a common component of partial-to-partial matching pipelines across point clouds, images, and shapes~\cite{wang2019prnet,huang2021predator,xu2021omnet,ehm2024partial}. Depending on the task and application, different evaluation protocols are used. In point cloud registration, overlap estimates are primarily employed to filter inliers for transformation estimation and are therefore assessed using task-specific measures such as inlier ratios and related variants. In contrast, image segmentation and shape matching focus on dense region predictions and often evaluate overlap prediction using Intersection-over-Union (IoU), due to its direct geometric interpretation and alignment with visual overlap.
In partial-to-partial shape matching, overlap ratios can vary substantially across instances, ranging from near-zero to full overlap. In these cases, IoU scores can be influenced by the overlap ratio, making comparisons across instances with different overlap ratios difficult. To better analyze overlap prediction quality across varying overlap ratios, we additionally include balanced accuracy, a metric commonly used in imbalanced binary classification, for complementary insights.

\noindent\textbf{Software Frameworks.}  
In recent years, there have been growing efforts to provide practical software support for functional map methods.
GeomFuM~\cite{vigano2025geomfum} provides a comprehensive Python package for adoption of functional maps in geometric deep learning pipelines, supporting generalized applications across tasks beyond shape matching. 
pyFM~\cite{pyfmaps2025} focuses on clear, algorithmic implementations of classical functional map pipelines, prioritizing readability, explicit formulations, and ease of modification, but not designed for deep learning pipelines.
In contrast, our codebase focuses explicitly on deep shape matching workflows, following the design of \cite{cao_unsupervised_2023}, with an emphasis on standardized evaluation, modular components, and efficient implementations.%

\section{Method}
\label{sec:method}

Spectral methods are a fundamental pillar of 3D shape correspondence; for an overview, we refer the reader to a recent survey \cite{zhuravlev2026nonrigid}.
In the following, we introduce an accelerated functional map solver that significantly improves computational efficiency while remaining mathematically equivalent to existing formulations.

\subsection{Batched Functional Map Solver}
We begin by expressing the computed feature descriptors in the respective spectral bases, denoted by $\Phi_{1}$ and $\Phi_{2}$. Let $\mathbf{F}$ and $\mathbf{G}$ denote the per-vertex feature descriptors extracted from the two shapes, respectively. This yields the spectral descriptors
\begin{equation}
\mathbf{A} = (\Phi_{1})^{\dagger} \mathbf{F},
\qquad
\mathbf{B} = (\Phi_{2})^{\dagger} \mathbf{G}
\end{equation}
where $(\cdot)^{\dagger}$ denotes the Moore--Penrose pseudo-inverse. This step shifts the correspondence estimation problem from the spatial domain to the spectral domain.

Following the seminal framework introduced by Ovsjanikov et al.~\cite{ovsjanikov2012functional} and widely adopted in subsequent work~\cite{ovsjanikov2017computing, attaiki2021dpfm, cao_unsupervised_2023, bastian2024hybrid,li_learning_2022,xie2025echomatch}, the functional map $\mathbf{C}$ is obtained by minimizing the energy
\begin{equation}
\min_{\mathbf{C}}
\;\big\Vert \mathbf{C}\mathbf{A} - \mathbf{B} \big\Vert^2
\;+\;
\lambda \big\Vert \mathbf{C}\Delta_{1} - \Delta_{2}\mathbf{C} \big\Vert^2,
\label{eq:fmap_lap}
\end{equation}
where $\Delta_{1}$ and $\Delta_{2}$ are the Laplace--Beltrami operators on the two shapes, and $\lambda$ is a scalar regularization parameter. The second term enforces Laplacian commutativity to promote isometry.

Throughout the literature, several strategies have been proposed to compute $\mathbf{C}$ in learning-based settings. Early approaches~\cite{litany_deep_2017,halimi_unsupervised_2019,roufosse2019unsupervised} optimized only the data term $\|\mathbf{C}\mathbf{A}-\mathbf{B}\|^2$, omitting regularization. While simple, this formulation is highly sensitive to the conditioning of $\mathbf{A}$, especially during early training, and can lead to numerical instability. A more complete solution can be obtained by vectorizing $\mathbf{C}$ and solving a large least-squares system~\cite{ovsjanikov2012functional,eisenberger2020smooth}, similarly as with non-orthogonal bases under the Hilbert--Schmidt norm~\cite{hartwig_elastic_2023,bastian2024hybrid}. However, these approaches result in a system of size $k^2\times k^2$, where $k$ is the spectral resolution, and are therefore only tractable for small $k$.

\paragraph{Row-wise.}
The solution most commonly used in modern deep functional map pipelines was introduced in GeomFmaps~\cite{donati2020deep}. Noting that $\Delta_{1}$ and $\Delta_{2}$ are diagonal in their respective eigenbases, in the form as eigenvalues $\Lambda_{1}$ and $\Lambda_{2}$, the commutativity term admits the element-wise form
\begin{equation}
\big\Vert \mathbf{C}\Delta_{1} - \Delta_{2}\mathbf{C} \big\Vert_F^2
\;=\;
\sum_{i,j} (\Lambda_{2}(i) - \Lambda_{1}(j))^2\, C_{ij}^2.
\label{eq:fmap_lap_weighted}
\end{equation}
This corresponds to an entry-wise weighted Frobenius norm.

Minimizing the resulting quadratic objective leads to the linear system
\begin{equation}
\mathbf{C}\mathbf{A}\mathbf{A}^T
\;+\;
\lambda\,(\mathbf{M}\odot\mathbf{C})
\;=\;
\mathbf{B}\mathbf{A}^T,
\end{equation}
where $\odot$ denotes element-wise multiplication and the penalty mask
$\mathbf{M}\in\mathbb{R}^{k\times k}$ is defined by
$\mathbf{M}(i,j)=(\Lambda_{2}(i)-\Lambda_{1}(j))^2$.
This mask has been extensively studied~\cite{ren2019structured} and can be replaced by more structured alternatives, such as the resolvent mask~\cite{ren2019structured}, used in DPFM~\cite{attaiki2021dpfm} and many subsequent works~\cite{cao_unsupervised_2023, bastian2024hybrid,li_learning_2022,xie2025echomatch}.

Crucially, this system decouples row-wise. For each row $\mathbf{c}_i$ of $\mathbf{C}$, we obtain an independent $k\times k$ linear system
\begin{equation}
\big(\mathbf{A}\mathbf{A}^T + \lambda\,\mathrm{diag}(\mathbf{m}_i)\big)\,\mathbf{c}_i
\;=\;
(\mathbf{B}\mathbf{A}^T)_i
\label{eq:lapreg}
\end{equation}
with solution
\begin{equation}
\mathbf{c}_i
=
\operatorname{solve}\!\left(
\mathbf{A}\mathbf{A}^T + \lambda\,\mathrm{diag}(\mathbf{m}_i),
\;(\mathbf{B}\mathbf{A}^T)_i
\right)
\end{equation}
where $i$ denotes the $i$-th row of the respective matrix.
The full functional map is then assembled row by row as
\begin{equation}
\mathbf{C}
=
\begin{bmatrix}
\mathbf{c}_1^T \\
\mathbf{c}_2^T \\
\vdots \\
\mathbf{c}_k^T
\end{bmatrix}
\end{equation}
In total, this requires solving $k$ distinct $k\times k$ systems, typically implemented in a loop~\cite{donati2020deep,attaiki2021dpfm,cao_unsupervised_2023}. While conceptually simple, the loop-based formulation becomes a computational bottleneck as the spectral resolution $k$ increases.

\paragraph{Batched Row-wise.}
We observe that these $k$ independent systems can be solved fully in parallel by treating the row index as a batch dimension.
We therefore reformulate the computation as a single batched tensor solve:
\begin{equation}
\mathbf{C}
=
\operatorname{solve}\Big(
\underbrace{\mathbf{A}\mathbf{A}^T}_{[b,1,k,k]}
+
\lambda\,\underbrace{\mathrm{diagembed}(\mathbf{M})}_{[b,k,k,k]},
\;
\underbrace{\mathbf{B}\mathbf{A}^T}_{[b,k,k,1]}
\Big)
\end{equation}
where $b$ denotes the batch size and $k$ the spectral resolution.
$\mathrm{diagembed}(\mathbf{M})$ expands each row $\mathbf{m}_i$ into a diagonal matrix along a batch dimension, producing a tensor of shape $[b,k,k,k]$. After broadcasting and summation, the system can be written compactly as
\begin{equation}
\mathbf{C}
=
\operatorname{solve}\big(
\underbrace{\mathrm{Lhs}}_{[b,k,k,k]},
\;
\underbrace{\mathrm{Rhs}}_{[b,k,k,1]}
\big)
\end{equation}
where the first two dimensions $(b,k)$ are treated as batch dimensions and the last two dimensions correspond to individual linear systems.
The operator $\operatorname{solve}(\cdot)$ denotes a batched linear solver, implemented using \texttt{torch.linalg.solve}, which computes all $k$ solutions simultaneously within a single kernel call. The resulting tensor is the functional map $\mathbf{C}\in\mathbb{R}^{b\times k\times k}$.
This batched formulation is mathematically equivalent to the standard row-wise solver, but substantially more efficient in practice. At the commonly used spectral resolution of $k=200$, it achieves a $33\times$ speedup over the loop-based implementation, with no numerical differences. Figure~\ref{fig:spectral_runtime} illustrates the runtime across spectral resolutions. A small trade-off is memory: the batched formulation materializes a $[b,k,k,k]$ tensor,
a factor $k$ increase over the $[b,k,k]$ working memory of the loop version.
In practice this overhead is modest even at $k{=}300$, with no more than $200\,\text{MB}$ additional allocation per map.

\begin{figure}[t]

\begin{codecard}{python}
C = []
for i in range(k):
    diag_mi = torch.diag(M[i])
    lhs = AAt + lmbda * diag_mi
    rhs = BAt[i]
    c_i = torch.linalg.solve(lhs, rhs)
    C.append(c_i)
C = torch.cat(C)
\end{codecard}
\vspace{-12pt}
\begin{center}
{\small Previous sequential solve~\cite{donati2020deep, attaiki2021dpfm, cao_unsupervised_2023, bastian2024hybrid,li_learning_2022,xie2025echomatch}}
\end{center}
\vspace{-12pt}

\begin{codecard}{python}
Diag_M = torch.diag_embed(M)
Lhs = AAt + lmbda * Diag_M
Rhs = BAt
C = torch.linalg.solve(Lhs, Rhs)
\end{codecard}
\vspace{-12pt}
\begin{center}
{\small Our batched solve}
\end{center}
\vspace{-12pt}
\caption{\textbf{Pseudocode comparison of functional map solvers.} 
Previous implementations~\cite{donati2020deep, attaiki2021dpfm, cao_unsupervised_2023, bastian2024hybrid,li_learning_2022,xie2025echomatch} solve each system sequentially, whereas our formulation performs the solves in a batch.}
\vspace{-4mm}
\label{fig:solver}
\end{figure}

\subsection{Learned Scaling and Rotation in DiffusionNet}
We document a subtle difference between two commonly used implementations of spatial gradient features in DiffusionNet based deep shape matching pipelines~\cite{sharp2022diffusionnet, attaiki2021dpfm, xie2025echomatch,li_learning_2022,donati2022deep,zeng2025coe,attaiki2023understanding,cao_unsupervised_2023, bastian2024hybrid,vigano2025geomfum,cao2024spectral,cao2022unsupervised,cao_self-supervised_2023,ehm2024partial,roetzer2024spidermatch}. These variants parameterize distinct families of linear transformations on tangent plane gradients.

The original DiffusionNet~\cite{sharp2022diffusionnet} leverages spatial gradients of per-vertex signals to construct expressive, orientation-aware features. Given scalar vertex features with $D$ channels, $F \in \mathbb{R}^{V \times D}$, their spatial gradients are computed as
\begin{equation}
Z = G(F) \in \mathbb{C}^{V \times D}
\end{equation}
where $G \in \mathbb{C}^{V \times V}$ is a precomputed sparse gradient operator that maps real-valued vertex functions to complex-valued tangent vectors expressed in an arbitrary local reference frame. While the choice of reference frame is arbitrary, inner products of these gradients are invariant and thus well-defined.

\begin{figure}[t]
    \centering
    \vspace{-4.0mm}
    \begin{minipage}[t]{0.50\columnwidth}
        \centering
        \includegraphics[width=\linewidth]{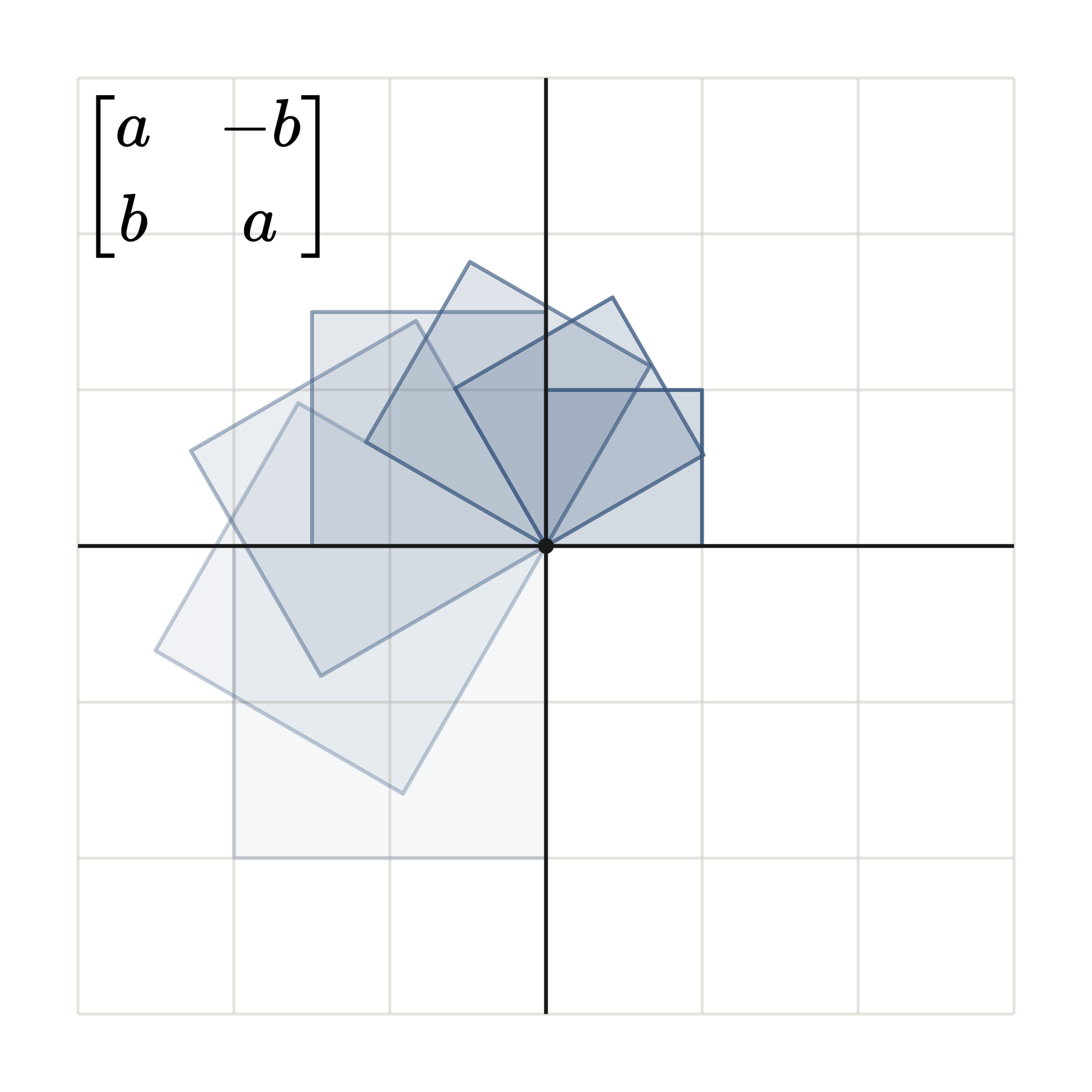}
        \vspace{-1.cm}
        \caption*{(A)}
        \vspace{-0.5cm}
    \end{minipage}\hfill
    \begin{minipage}[t]{0.50\columnwidth}
        \centering
        \includegraphics[width=\linewidth]{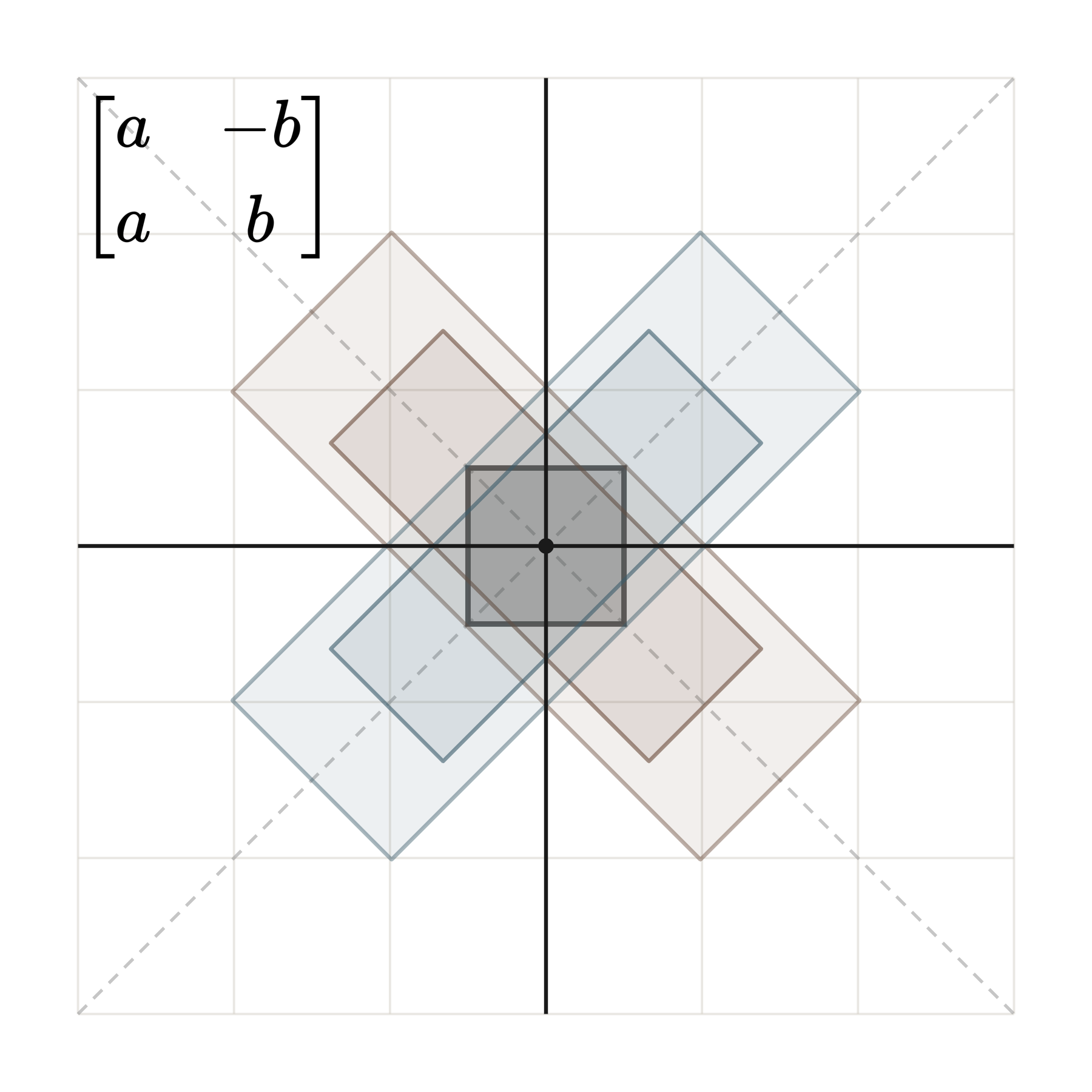}
        \vspace{-1.cm}
        \caption*{(B)}
        \vspace{-0.5cm}
    \end{minipage}
    \caption{\textbf{Illustration of two linear transform families in $\mathbb{R}^2$.} Left: Arbitrary rotation and isotropic scaling, used by~\cite{sharp2022diffusionnet, attaiki2021dpfm, xie2025echomatch,li_learning_2022,donati2022deep,zeng2025coe,attaiki2023understanding}; Right: fixed rotation at 45 degree angle and arbitrary anisotropic scaling used by~\cite{cao_unsupervised_2023, bastian2024hybrid,vigano2025geomfum,cao2024spectral,cao2022unsupervised,cao_self-supervised_2023,ehm2024partial,roetzer2024spidermatch}.}
    \label{fig:transformation}
\vspace{-4.0mm}
\end{figure}

The gradient features are processed independently at each vertex across channels. For a fixed vertex, we denote the stacked complex-valued gradients by $z \in \mathbb{C}^{D}$, representing $D$ local 2D tangent vectors, and aim to produce real-valued features $g \in \mathbb{R}^{D}$. DiffusionNet achieves this via
\begin{equation}
g = \mathrm{Re}\!\left( \bar{z} \odot A z \right)
\end{equation}
where $A \in \mathbb{C}^{D \times D}$ is a learned complex-valued matrix, $\odot$ denotes element-wise multiplication, and $\mathrm{Re}(\cdot)$ extracts the real part. The use of complex arithmetic provides a compact notation for learned rotations and scalings of 2D tangent vectors.

Writing $A = A_{\mathrm{re}} + i\, A_{\mathrm{im}}$ and $z = x + i\, y$ with $x, y \in \mathbb{R}^{D}$, the matrix--vector product $Az$ decomposes into real and imaginary parts as
\begin{equation*}
(Az)_{\mathrm{re}} = A_{\mathrm{re}}\, x - A_{\mathrm{im}}\, y, \qquad
(Az)_{\mathrm{im}} = A_{\mathrm{re}}\, y + A_{\mathrm{im}}\, x
\end{equation*}
and the full operation becomes
\begin{equation}
g \;=\; x \odot (Az)_{\mathrm{re}} \;+\; y \odot (Az)_{\mathrm{im}}
\end{equation}
which corresponds directly to the implementation shown in Fig.~\ref{fig:diffusionnet_variants}(A). Examining the $i$-th channel of the output $g$ and writing $\mathbf{v}_j = (x_j,\, y_j)^\top$ for the tangent vector of channel $j$, we obtain
\begin{equation}
g_i
=
\sum_{j=1}^{D}
\mathbf{v}_i^\top
\begin{bmatrix}
a & -b \\
b & \phantom{-}a
\end{bmatrix}
\mathbf{v}_j
\end{equation}
where $a = A_{\mathrm{re}}(i,j)$ and $b = A_{\mathrm{im}}(i,j)$. The $2\times2$ block
\(
\begin{bmatrix}
a & -b \\ b & a
\end{bmatrix}
\)
represents an arbitrary learned rotation combined with isotropic scaling in the tangent plane. This formulation enables the network to learn directional filters sensitive to local orientation, extending beyond purely radial responses.

A different DiffusionNet implementation, used by~\cite{cao_unsupervised_2023, bastian2024hybrid,vigano2025geomfum,cao2024spectral,cao2022unsupervised,cao_self-supervised_2023}, differs only in the imaginary part of the transformed gradient, see Fig.~\ref{fig:diffusionnet_variants}(B):
\begin{equation*}
(Az)_{\mathrm{re}} = A_{\mathrm{re}}\, x - A_{\mathrm{im}}\, y, \qquad
(Az)_{\mathrm{im}} = A_{\mathrm{re}}\, x + A_{\mathrm{im}}\, y
\end{equation*}

The per-channel output of $g$ becomes
\begin{equation}
g_i
=
\sum_{j=1}^{D}
\mathbf{v}_i^\top
\begin{bmatrix}
a & -b \\
a & \phantom{-}b
\end{bmatrix}
\mathbf{v}_j
\end{equation}
where
\(
\begin{bmatrix}
a & -b \\ 
a & b
\end{bmatrix}
\)
corresponds to a distinct family of linear transformations on 2D tangent vectors. In contrast to the original formulation, the learned rotations are fixed at $\pm45^\circ$ while allowing arbitrary anisotropic scalings.

\begin{figure}[t]
\vspace{-2.0mm}
\begin{codecard}{python}
Az_re = A_re(z[...,0]) - A_im(z[...,1])
Az_im = A_re(z[...,|\textbf{\textcolor{darkgreen}{1}}|]) + A_im(z[...,|\textbf{\textcolor{darkgreen}{0}}|])
g = z[...,0] * Az_re + z[...,1] * Az_im
\end{codecard}
\vspace{-12pt}
\begin{center}
    {\small Implementation (A) used by~\cite{sharp2022diffusionnet, attaiki2021dpfm, xie2025echomatch,li_learning_2022,donati2022deep,zeng2025coe,attaiki2023understanding}}
\end{center}
\vspace{-12pt}
\begin{codecard}{python}
Az_re = A_re(z[...,0]) - A_im(z[...,1])
Az_im = A_re(z[...,|\textbf{\textcolor{darkgreen}{0}}|]) + A_im(z[...,|\textbf{\textcolor{darkgreen}{1}}|])
g = z[...,0] * Az_re + z[...,1] * Az_im
\end{codecard}
\vspace{-12pt}
\begin{center}
    {\small Implementation (B) used by~\cite{cao_unsupervised_2023, bastian2024hybrid,vigano2025geomfum,cao2024spectral,cao2022unsupervised,cao_self-supervised_2023,ehm2024partial,roetzer2024spidermatch}}
\end{center}
\vspace{-10pt}
\caption{\textbf{Two implementations of spatial gradient features in DiffusionNet.} 
The variants differ only in the indexing used.}
\label{fig:diffusionnet_variants}
\vspace{-4.0mm}
\end{figure}

Figure~\ref{fig:transformation} illustrates the two transformation families.
As the two transformation families span different subspaces of linear operators on the tangent plane, they exhibit complementary behavior and can be advantageous in different scenarios, as demonstrated in our experiments.

\subsection{Overlap Prediction Metrics}

Predicting which regions of two shapes overlap is a central component of modern partial-to-partial correspondence pipelines~\cite{wang2019prnet,huang2021predator,xu2021omnet,ehm2024partial}.  
Overlap prediction appears across multiple domains, and the choice of evaluation metric has been guided by task-specific objectives.
In point cloud registration, overlap estimates are primarily used to filter inliers for downstream transformation estimation.  
As a result, evaluation often focuses on precision like measures that reflect the reliability of predicted correspondences, rather than region level agreement.  
In image segmentation, overlap prediction is closely aligned with the final task objective, and metrics such as Intersection-over-Union (IoU) are widely adopted due to their intuitive geometric interpretation and direct tie to visual overlap.
In shape matching, IoU has similarly become a standard metric for evaluating overlap prediction, as it directly measures agreement between predicted and ground-truth overlapping regions.

Our motivation for revisiting overlap evaluation metrics comes from the fact that, in partial-to-partial shape matching, the overlap ratio $r \in [0,1]$ can vary dramatically across shape pairs, ranging from near-complete overlap to extremely small intersections.  
This variability introduces a wide spectrum of class imbalance.

Under such conditions, commonly used metrics, including Accuracy, IoU, and F1, correlate largely with the overlap ratio $r$.  
As a result, metric values may reflect more differences in overlap ratio and predictive biases than differences in predictive accuracy.

To illustrate this effect, we analyze three degenerate predictors:  
(i) Zeros: predict all negatives,
(ii) Ones: predict all positives,
(iii) Random: each vertex is predicted positive randomly with probability $0.5$.

\begin{figure*}[t]
  \centering
  \vspace{-6.0mm}
  \setlength{\tabcolsep}{2pt}
  \begin{tabular}{cccc}
      \includegraphics[width=0.24\linewidth]{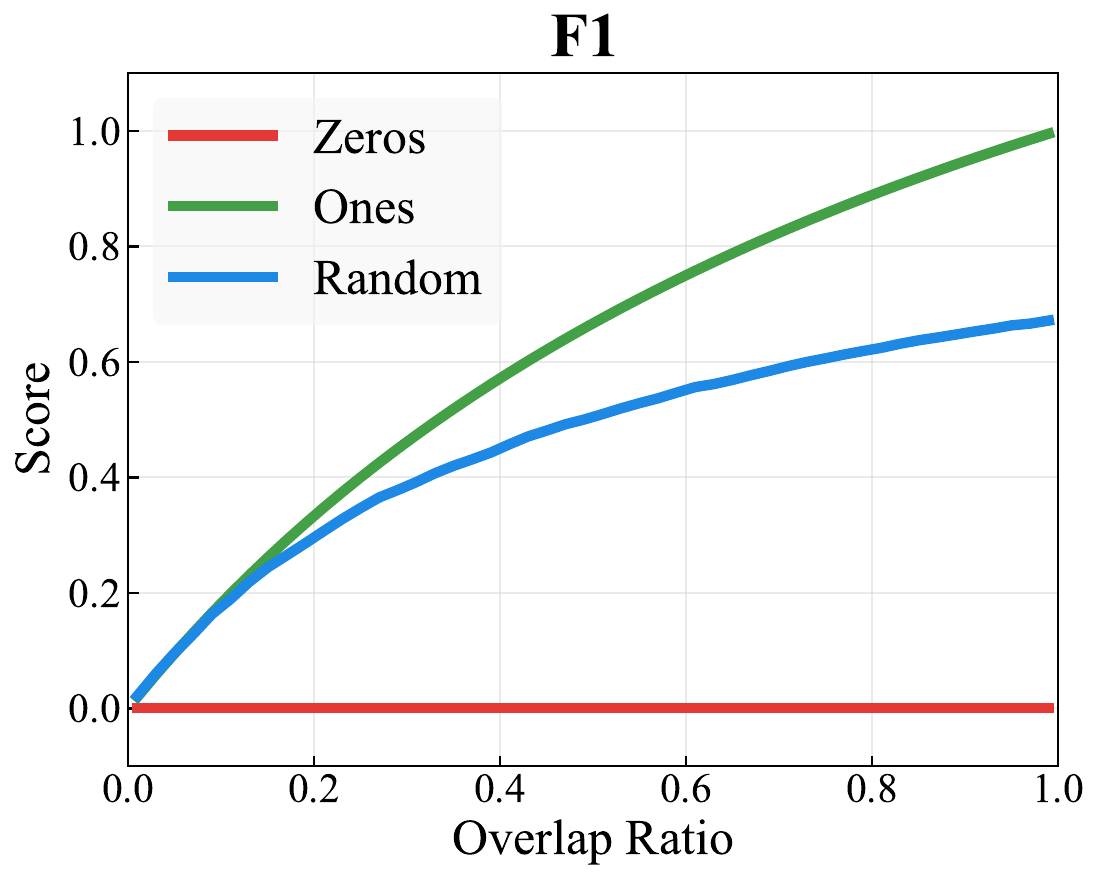} &
    \includegraphics[width=0.24\linewidth]{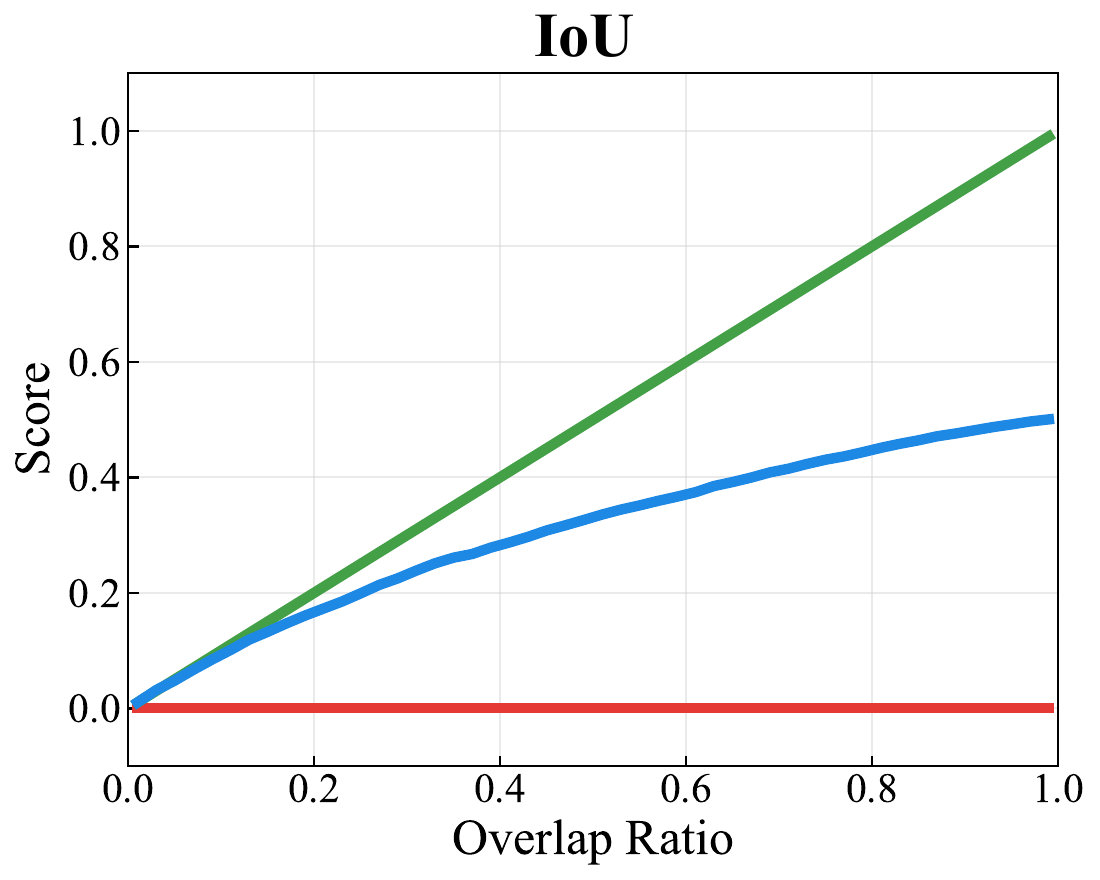} &
    \includegraphics[width=0.24\linewidth]{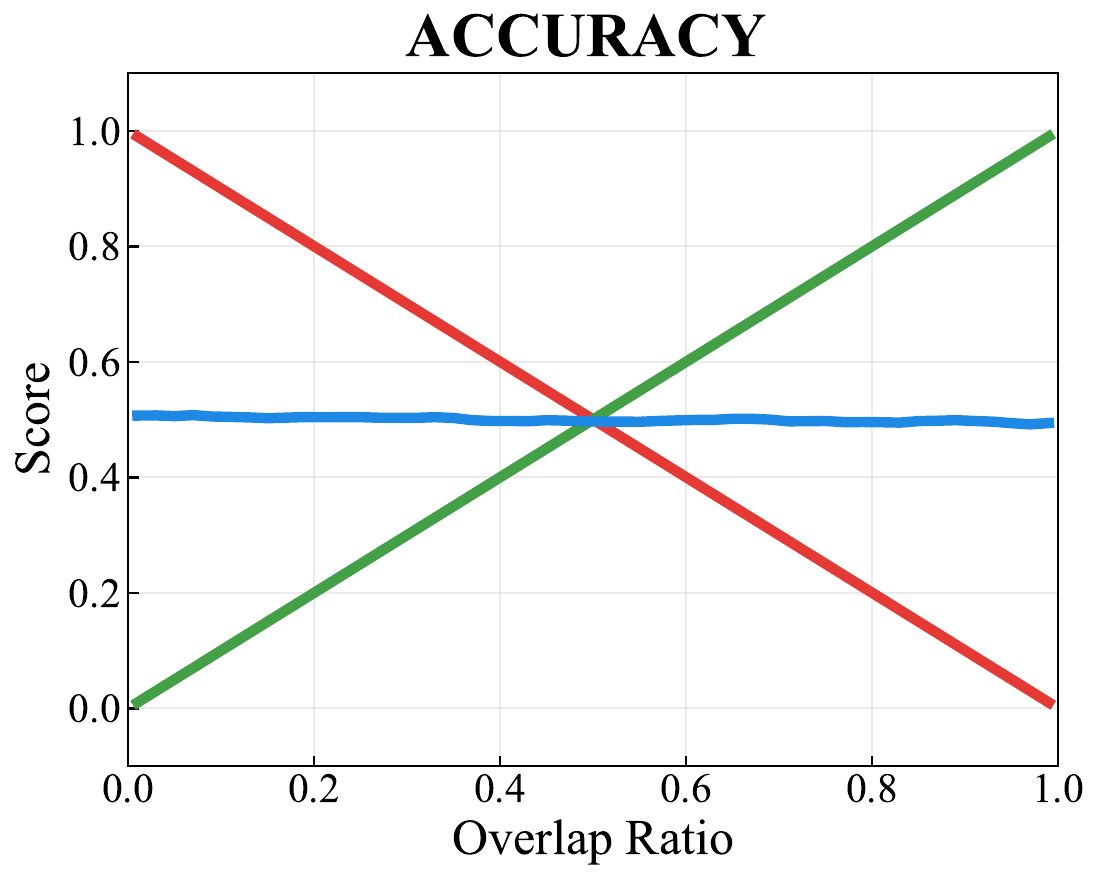} &
    \includegraphics[width=0.24\linewidth]{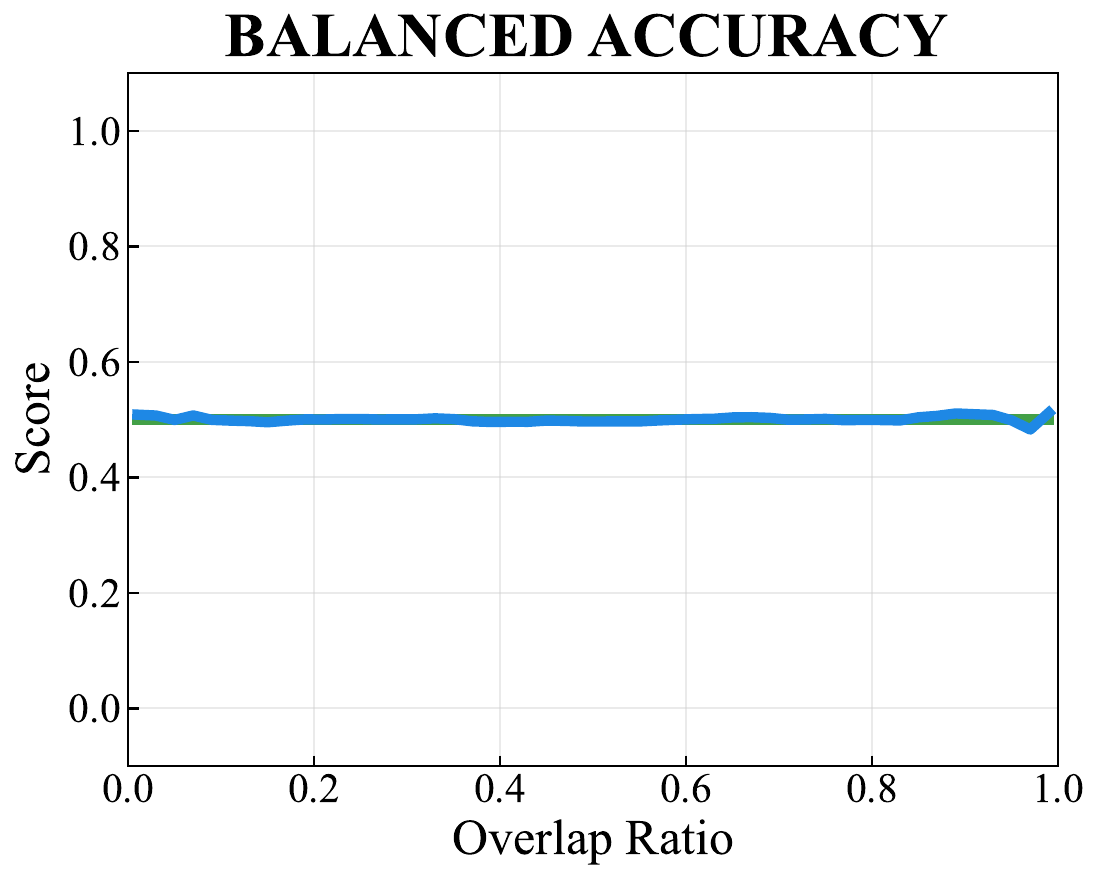}\\
    \includegraphics[width=0.24\linewidth]{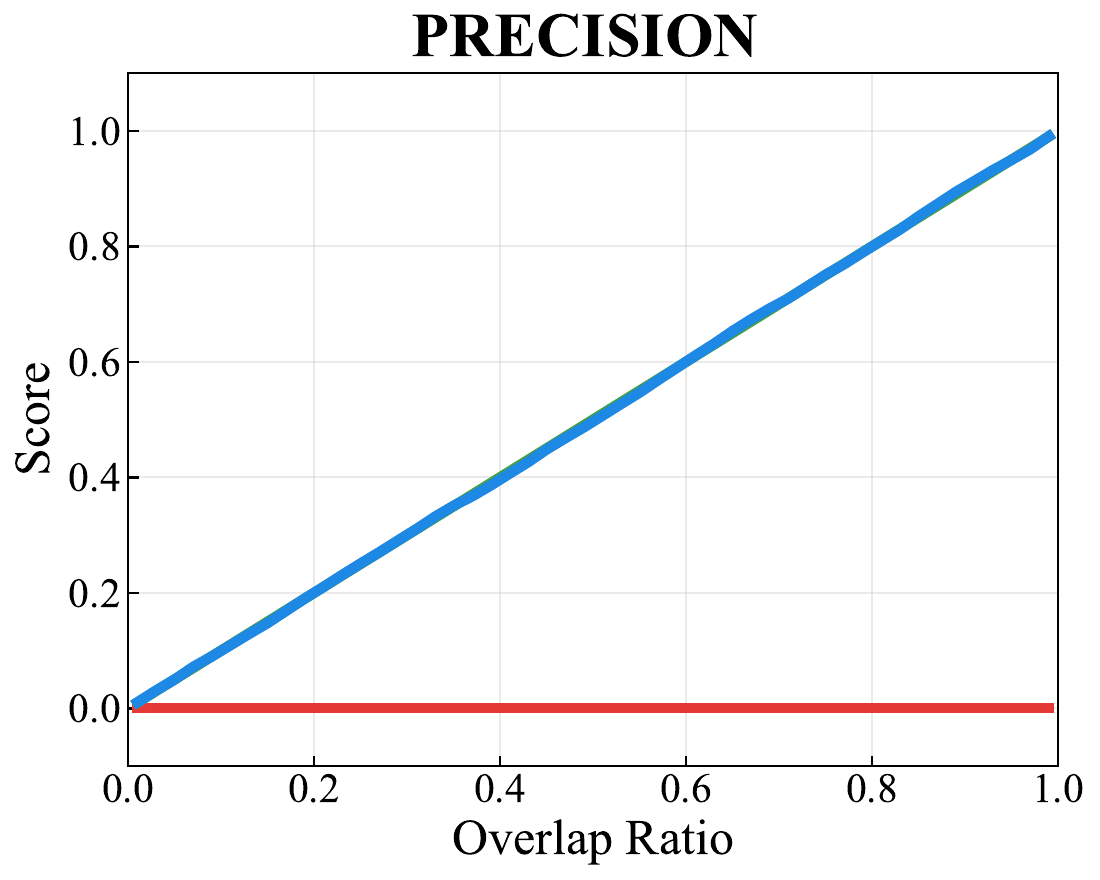} &
    \includegraphics[width=0.24\linewidth]{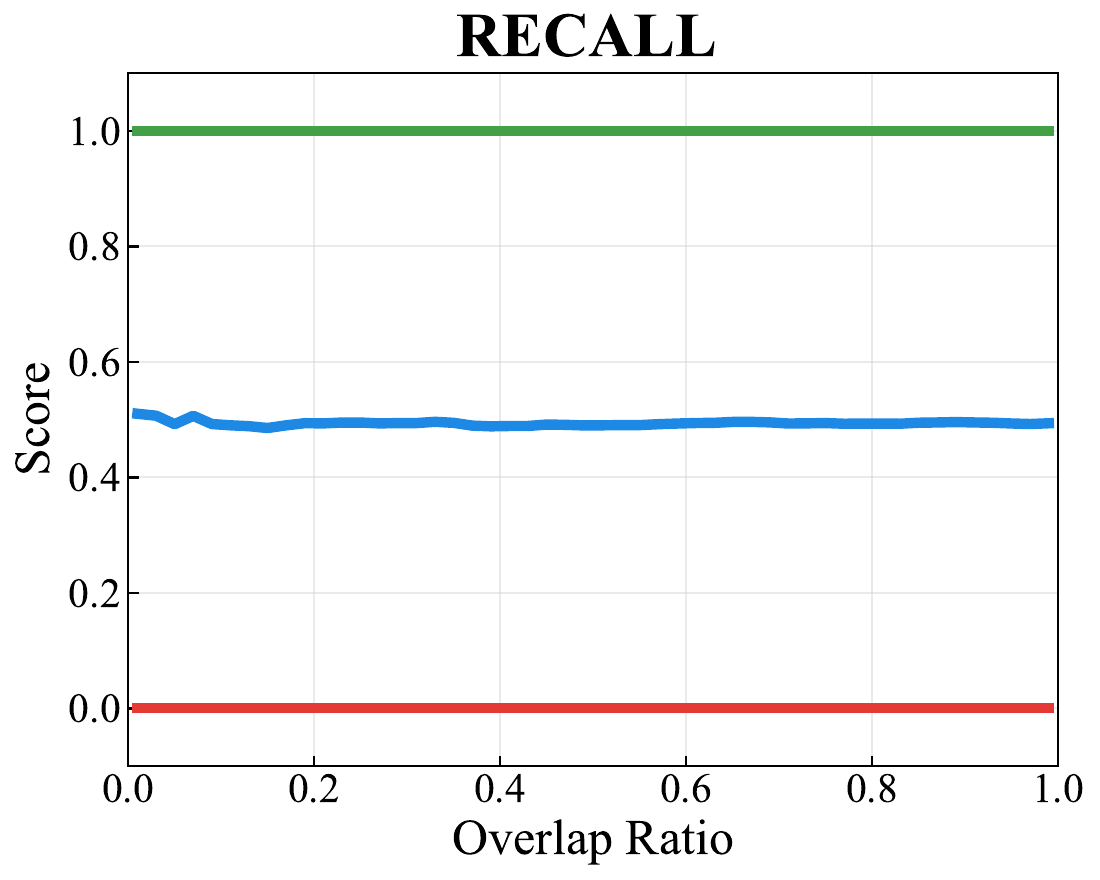} &
    \includegraphics[width=0.24\linewidth]{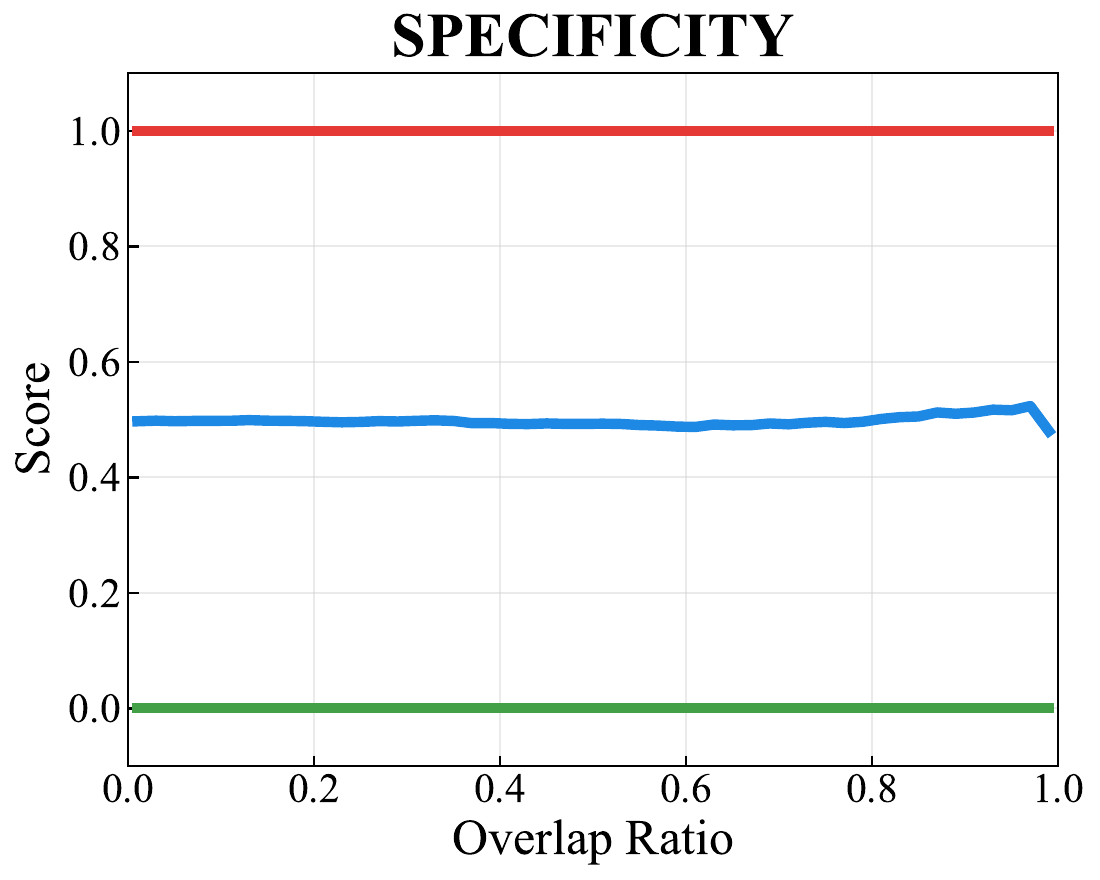} &
    \includegraphics[width=0.24\linewidth]{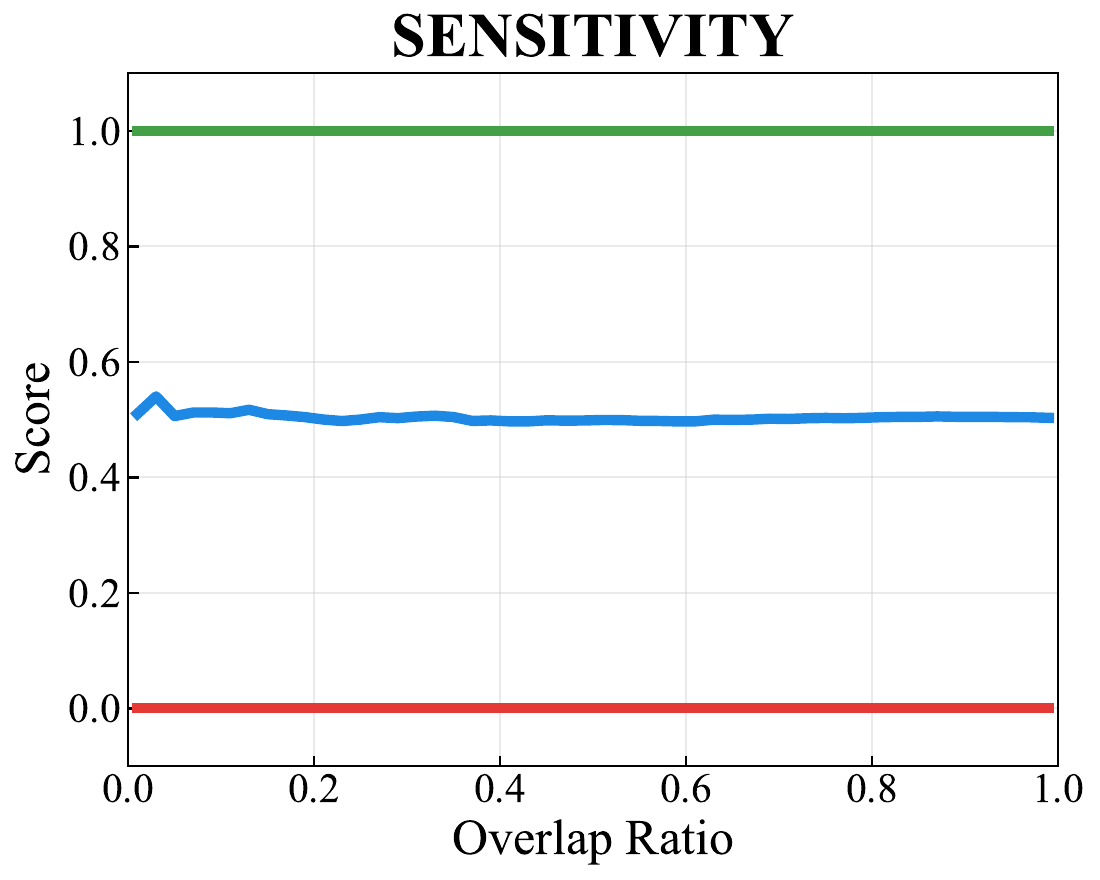} \\
  \end{tabular}
  \vspace{-1.0mm}
  \caption{\textbf{Metric behavior vs.\ overlap ratio.}
  X–axis: fraction of truly overlapping vertices $r\in[0,1]$; Y–axis: metric value.
  We simulate three degenerate predictors: \textcolor{red!0!black}{\emph{Zeros}} (predict no overlap), 
  \textcolor{green!0!black}{\emph{Ones}} (predict all overlap), and \textcolor{blue!0!black}{\emph{Random}} (each vertex predict overlap with prob.\ $0.5$).
  Most metrics are \emph{biased by the class prior} $r$ (e.g., Accuracy, Precision, IoU, F1),
  whereas \emph{Balanced Accuracy} remains constant at $0.5$ for all three baselines, reflecting predictive performance under class imbalance.} %
  \label{fig:overlap-metric-bias}
  \vspace{-4.0mm}
\end{figure*}

Writing TP/FP/FN/TN in expectation for a total region of size $N$:

\begin{equation*}
\begin{aligned}
Zeros\textbf{: } & \text{TP}=0,\;\text{FP}=0,\;\text{FN}=rN,\;\text{TN}=(1-r)N,\\
Ones\textbf{: }  & \text{TP}=rN,\;\text{FP}=(1-r)N,\;\text{FN}=0,\;\text{TN}=0,\\
Random\textbf{: }& \text{TP}=\tfrac12 rN,\;\text{FP}=\tfrac12 (1-r)N,\\
                 & \text{FN}=\tfrac12 rN,\;\text{TN}=\tfrac12 (1-r)N.
\end{aligned}
\end{equation*}

As shown in Fig.~\ref{fig:overlap-metric-bias}, most metrics vary systematically with $r$ even for these trivial baselines. Balanced Accuracy, defined as the average of sensitivity ($\text{TP}/(\text{TP}+\text{FN})$) and specificity ($\text{TN}/(\text{TN}+\text{FP})$), remains constant at $0.5$ for all three baselines regardless of class imbalance.

Balanced Accuracy is known to treat positive and negative classes symmetrically, which may not always align with application objectives; IoU remains a more intuitive choice when geometric region agreement is the primary concern.
We therefore include Balanced Accuracy as a complementary metric alongside IoU, and leave room for further task-driven overlap prediction metrics.

\section{Experiments}
We evaluate our contributions across a diverse set of benchmarks spanning near-isometric, non-isometric, topologically noisy, and partial matching settings, using a unified codebase with standardized training and evaluation pipelines.
For complete shape matching, FAUST~\cite{bogo2014faust}, SCAPE~\cite{anguelov2005scape}, and SHREC'19~\cite{melzi_shrec_2019} cover near-isometric human poses, while SMAL~\cite{Zuffi:CVPR:2017} and DT4D-H~\cite{li20214dcomplete,magnet2022smooth} target inter-class and highly non-isometric deformations.
TOPKIDS~\cite{lahner_shrec16_2016} tests robustness to topological noise via synthetic self-intersecting shapes.
For partial matching, SHREC'16~\cite{cosmo2016shrec} provides CUTS and HOLES subsets; CP2P24~\cite{attaiki2021dpfm,ehm2024partial} and PSMAL~\cite{ehm2024partial} extend this to partial-to-partial settings, and BeCoS~\cite{becos} offers a larger and more challenging benchmark.
All experiments are performed on an Intel Core i9-9900K CPU and NVIDIA RTX 2080 Ti GPU.

\begin{table*}[t]
\setlength{\tabcolsep}{6pt}
\centering
\small
\caption{\textbf{Per-iteration runtime and memory comparison} between the original implementations of multiple shape matching methods and using our batched solver on the same hardware. 
Additional GPU memory usage of our implementation is also reported. Original GPU memory consumption of the full learning pipelines ranges from approximately 1.5 to 8.4\,GB across methods.}
\label{tab:runtime}
\begin{tabular}{@{}lcccccc@{}}
\toprule
Implementation & ULRSSM & Hybrid ULRSSM & AttentiveFmaps & AttentiveFmaps-Fast & DPFM & EchoMatch \\
\midrule
Original & 429\,ms & 555\,ms & 2050\,ms & 578\,ms & 115\,ms & 215\,ms \\
Ours & \textbf{82\,ms} & 363\,ms & 448\,ms &416\,ms & 95\,ms & 195\,ms \\
\midrule
Speedup ($\times$) & \textbf{5.23}  & 1.53 & 4.57 &1.39 & 1.21 & 1.10 \\
\midrule
Memory ($+$) & 126\,MB (8.5\%) & 30\,MB (1.0\%) & 142\,MB (5.6\%) & 76\,MB (3.2\%) & 2\,MB (0.1\%) & 2\,MB (0.02\%) \\
\bottomrule
\end{tabular}
\vspace{-2mm}
\end{table*}

\paragraph{Runtime Improvements.}
We benchmark the proposed batched functional map solver against standard loop-based implementations, both in isolation across spectral resolutions and from within existing end-to-end shape matching pipelines.
As shown in Fig.~\ref{fig:spectral_runtime}, our batched solver achieves between $17\times$ and $33\times$ speedups over loop-based implementations across spectral resolutions of 30 to 300, with numerically equivalent results.
Table~\ref{tab:runtime} reports end-to-end per-iteration gains across multiple existing shape matching methods.
ULRSSM, the most widely adopted spectral shape matching method~\cite{cao_unsupervised_2023, ehm2023geometrically,ehm2024partial,roetzer2024spidermatch}, benefits most with a $5.2\times$ speedup, since its runtime is dominated by the functional map solver operating at a high spectral resolution of $k\!=\!200$.
AttentiveFmaps also achieves substantial gains of $4.6\times$ for the same reason: it solves multiple functional maps at multiple resolutions up to $k\!=\!200$.
Methods such as DPFM and EchoMatch see more modest speedups of $1.1$ to $1.2\times$, as they use a lower resolution of $k\!=\!50$ and their runtimes are dominated by other components: attention layers in DPFM, and dense point-map multiplications with an additional DiffusionNet in EchoMatch.
Overall, the magnitude of improvement scales directly with the fraction of runtime spent in the functional map solver and the spectral resolution used.
The additional GPU memory overhead remains modest across all methods, ranging from 2 to 142\,MB under 9\% of baseline consumption.

\begin{figure}[t]
\centering

\def\heightNI{2.6cm}

\setlength{\tabcolsep}{2pt}
\renewcommand{\arraystretch}{0.85}

\begin{tabular}{c c c c}

 & Source & (A) & (B) \\[2pt]

\rotatebox{90}{ \ \ \ DT4D-H} &
\includegraphics[height=\heightNI]{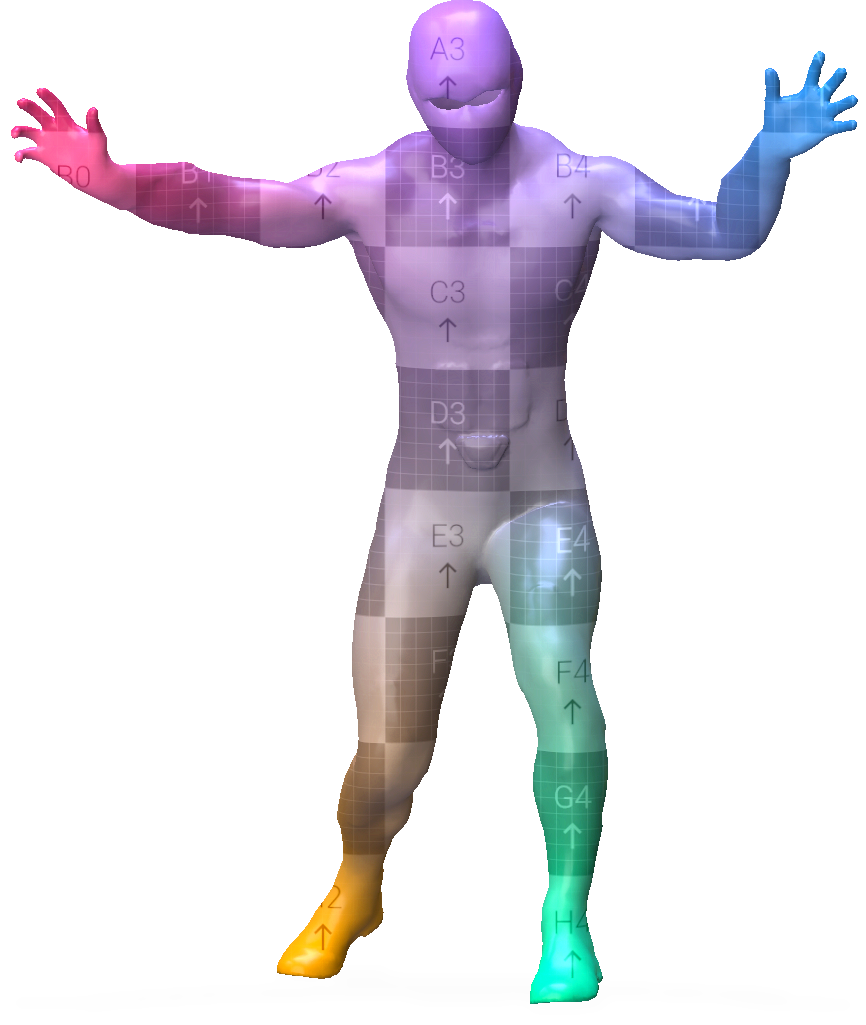} &
\includegraphics[height=\heightNI]{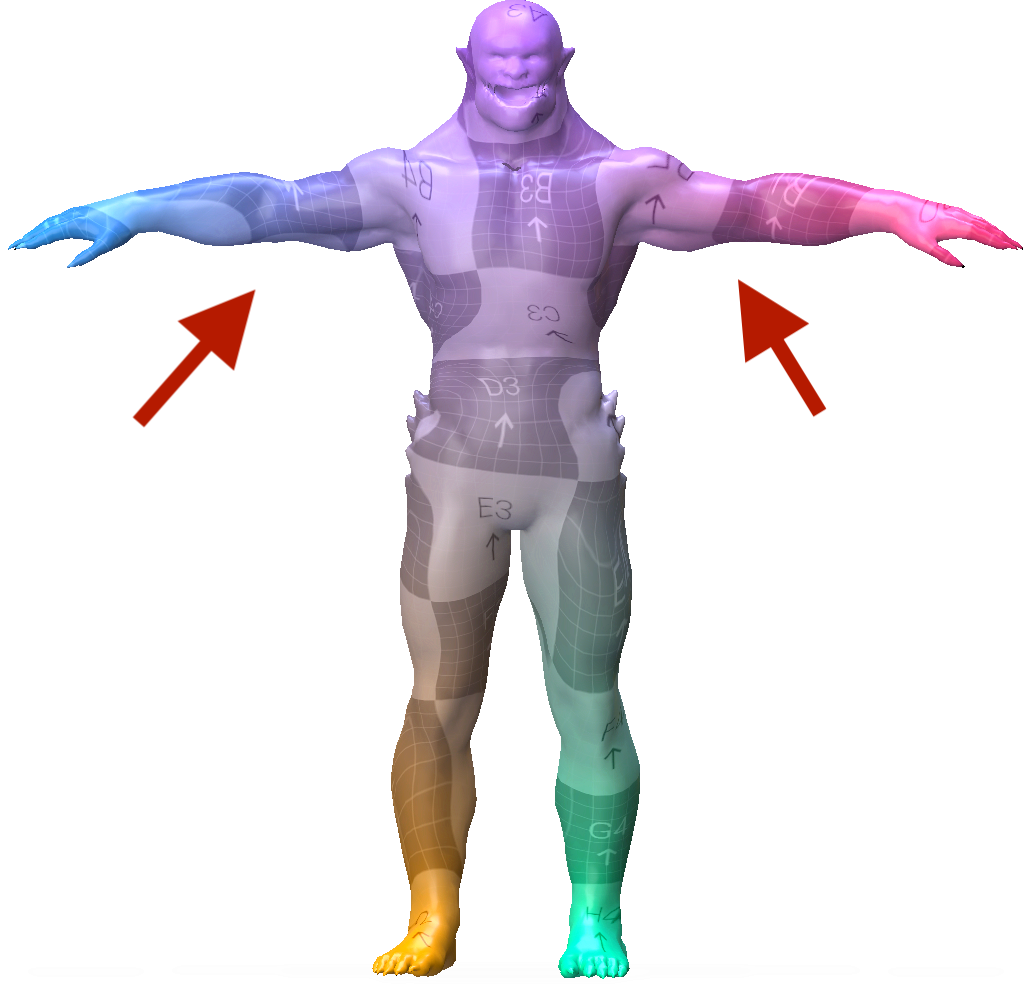} &
\includegraphics[height=\heightNI]{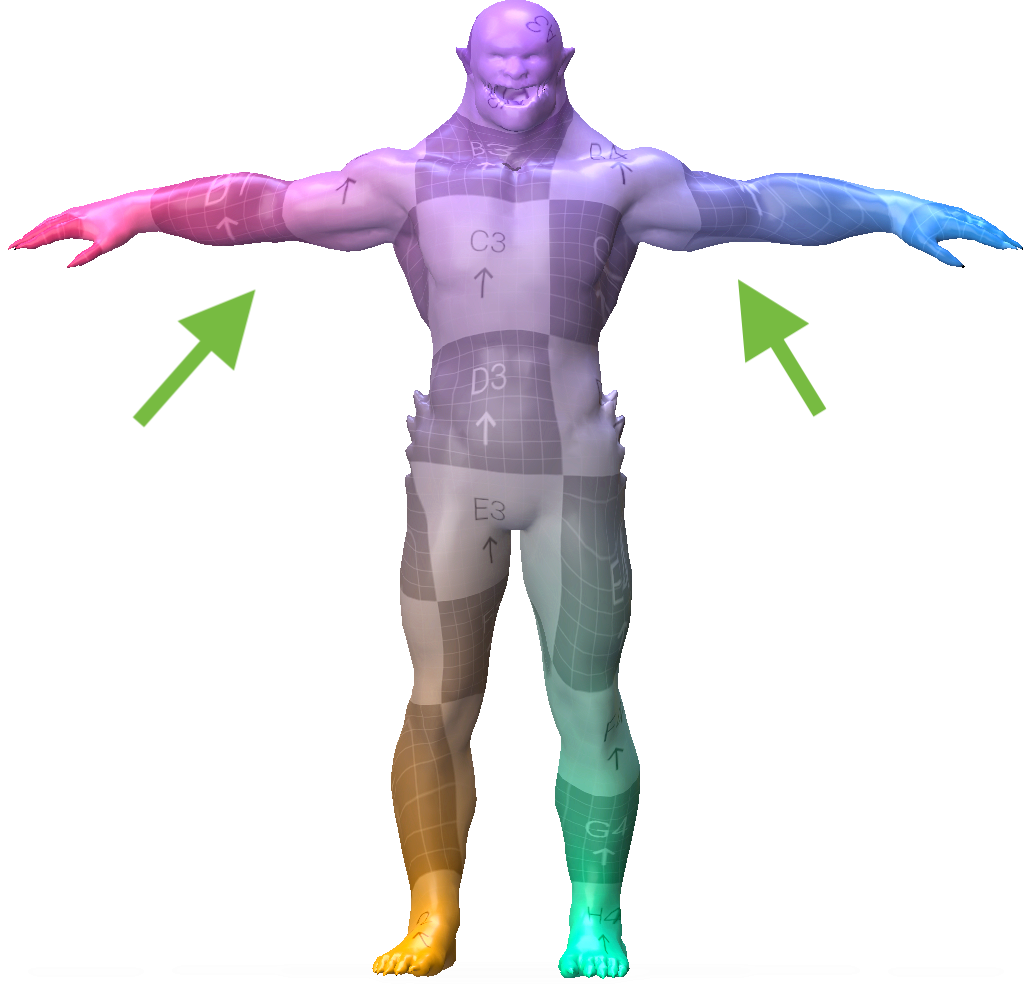} \\[0pt]

\rotatebox{90}{\ \ \ TOPKIDS} &
\includegraphics[height=\heightNI]{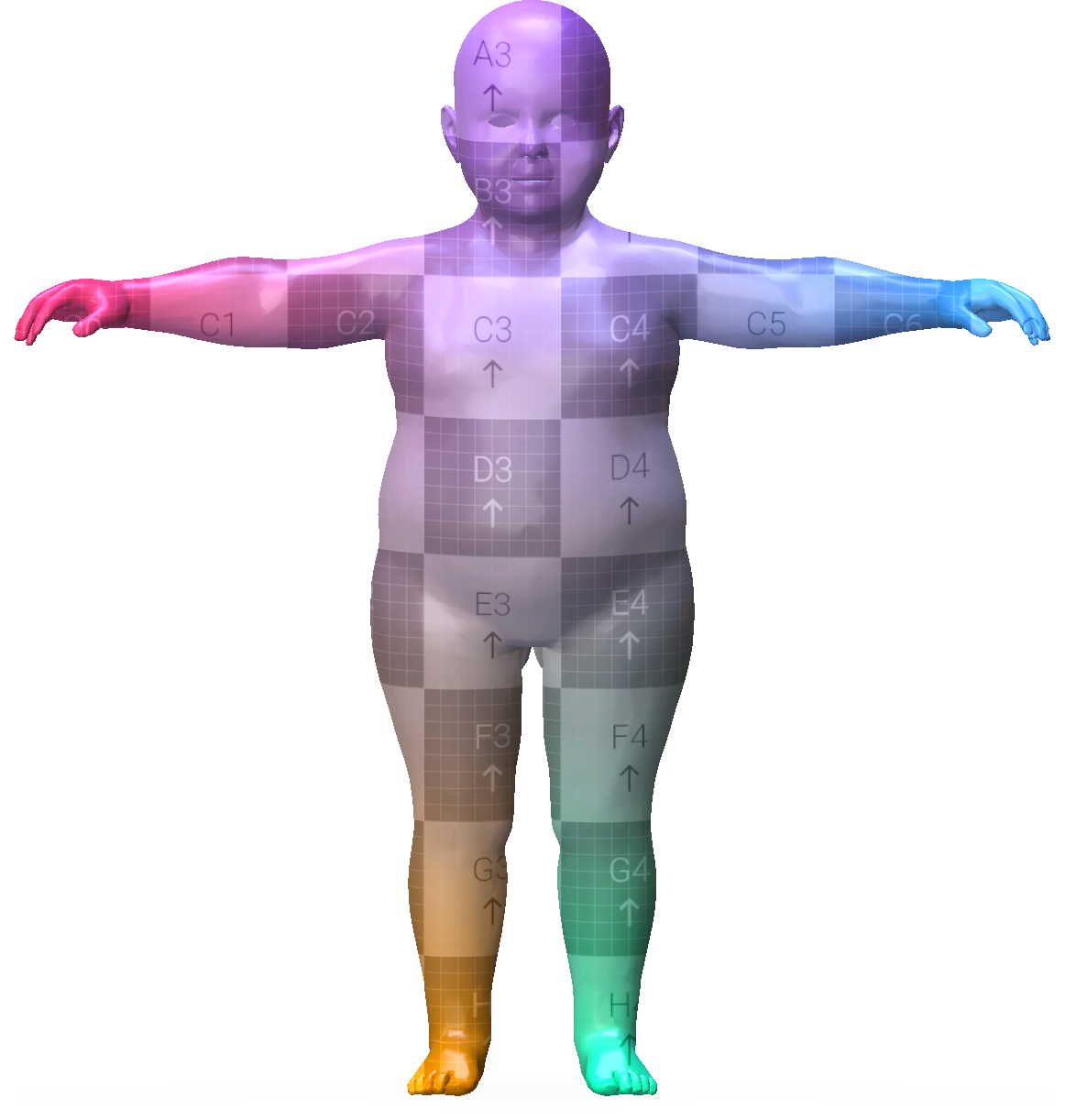} &
\includegraphics[height=\heightNI]{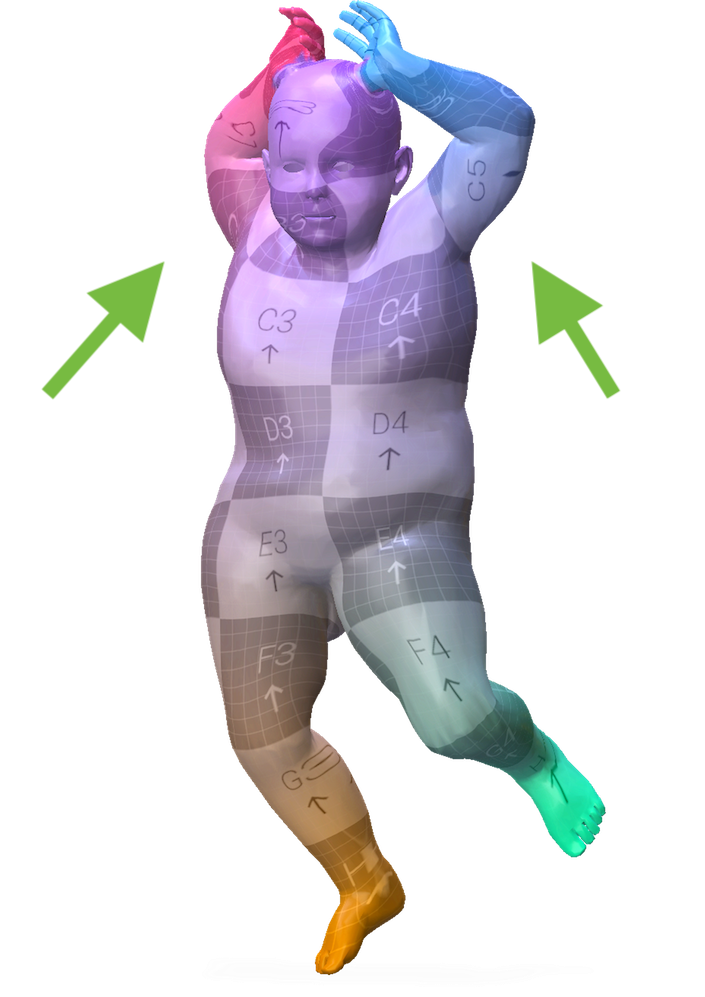} &
\includegraphics[height=\heightNI]{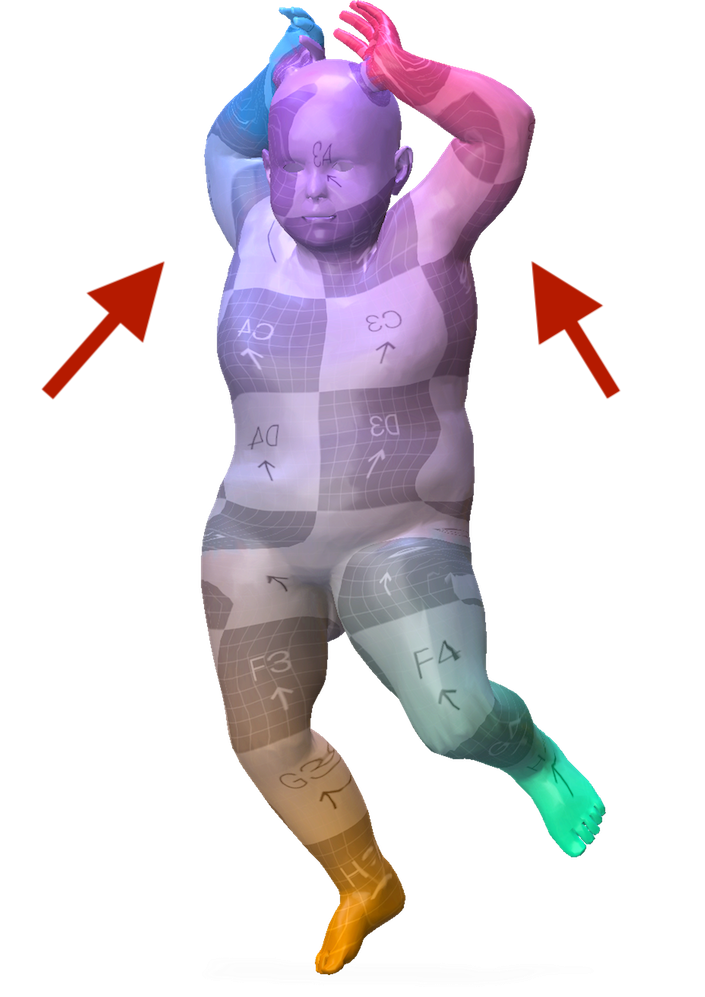}

\end{tabular}

\vspace{-2mm}

\caption{\textbf{Qualitative comparison} of DiffusionNet variants on DT4D-H and TOPKIDS using ULRSSM. Formulation A and B exhibit complementary strengths at non-isometry vs topological noise challenges.}
\label{fig:qualitative_cross_dataset}
\vspace{-4mm}
\end{figure}

\begin{table}[h!]
    \setlength{\tabcolsep}{3pt}
    \small
    \centering
    \caption{\textbf{Evaluation on non-isometric and topologically noisy datasets.}
    Average geodesic error ($\times 100$) across standard benchmarks.}
    \label{tab:gradrot_non_iso}
    \resizebox{\columnwidth}{!}{%
    \begin{tabular}{@{}cl|ccc|c@{}}
    \toprule
    & \multicolumn{1}{l|}{\multirow{2}{*}{\textbf{Geodesic Error ($\times$100)}}} 
    & \textbf{SMAL} & \multicolumn{2}{c|}{\textbf{DT4D-H}} & \textbf{TOPKIDS} \\
    & \multicolumn{1}{l|}{} & & \textbf{intra-class} & \textbf{inter-class} & \\
    \midrule

    & ULRSSM (B) &    3.9&  \textbf{0.9}&  \textbf{4.1}&  9.2\\

    & ULRSSM (A) &  3.9& 1.0 & 7.9 & \textbf{5.7}\\
    \midrule
    &Hybrid ULRSSM (B) &  3.3& 1.0& \textbf{3.5}& 4.6\\

    &Hybrid ULRSSM (A) & 3.3 & \textbf{0.9} & 3.9 & 4.6\\

    \bottomrule
    \end{tabular}
    }
    \vspace{-4mm}
\end{table}

\paragraph{DiffusionNet Variants.}
We study the effect of two commonly used DiffusionNet variants across a broad range of benchmarks.
We denote the original formulation with learned rotation and isotropic scaling as \emph{A}, and the alternative with a fixed $45^\circ$ rotation and learned anisotropic scaling as \emph{B}.
Both variants are susceptible to left--right symmetry flips, a known challenge for intrinsic methods.
On near-isometric humanoid benchmarks reported in Table~\ref{tab:near-iso}, both perform similarly, with \emph{A} showing a slight advantage, consistent with the invariance of inner products under rotation and isotropic scaling.
On strongly non-isometric shapes reported in Table~\ref{tab:gradrot_non_iso}, particularly DT4D-H where non-isometric deformations and bilateral symmetries co-occur as illustrated in Fig.~\ref{fig:qualitative_cross_dataset}, \emph{B} yields a clear advantage.
A plausible explanation is that anisotropic scaling provides a more expressive space of gradient transformations when symmetry cues are distorted by non-isometric stretching.
On TOPKIDS, which features near-isometric deformations under topological noise from self-intersecting meshes, \emph{A} performs more robustly.
In most partial matching settings reported in Tables~\ref{tab:gradrot_partial} and~\ref{tab:rotation_p2p}, \emph{A} is comparable or better, with less clear trends on the more challenging SHREC'16 HOLES and BeCoS benchmarks.
Overall, the two variants perform similarly in most settings but exhibit complementary strengths in specific scenarios.
The observed differences hint at a meaningful interaction between the choice of tangent-plane transformation and the nature of the deformations involved, though the evidence here is insufficient to draw firm conclusions.
We provide both implementations and empirical results and view deeper investigation into this interaction as an interesting direction for future work.

\begin{table*}[t]
\small
\setlength{\tabcolsep}{5pt}
    \centering
    \small
\caption{\textbf{Cross-dataset near-isometric evaluation on FAUST, SCAPE, and SHREC’19 datasets. }  
Average geodesic error ($\times 100$) across standard benchmarks.
}

        \resizebox{\textwidth}{!}{%
        \begin{tabular}{@{}lccccccccc@{}}
        \toprule
        \multicolumn{1}{l}{Train}  & \multicolumn{3}{c}{\textbf{FAUST}}   & \multicolumn{3}{c}{\textbf{SCAPE}}  & \multicolumn{3}{c}{\textbf{FAUST + SCAPE}} \\ \cmidrule(lr){2-4} \cmidrule(lr){5-7} \cmidrule(lr){8-10}
        \multicolumn{1}{l}{Test} & \multicolumn{1}{c}{\textbf{FAUST}} & \multicolumn{1}{c}{\textbf{SCAPE}} & \multicolumn{1}{c}{\textbf{SHREC'19}} & \multicolumn{1}{c}{\textbf{FAUST}} & \multicolumn{1}{c}{\textbf{SCAPE}} & \multicolumn{1}{c}{\textbf{SHREC'19}} & \multicolumn{1}{c}{\textbf{FAUST}} & \multicolumn{1}{c}{\textbf{SCAPE}} & \multicolumn{1}{c}{\textbf{SHREC'19}}
        \\ \midrule
        
        \multicolumn{1}{l}{ULRSSM (B)} & \multicolumn{1}{c}{1.6}  & \multicolumn{1}{c}{2.2} & \multicolumn{1}{c}{6.7}  & \multicolumn{1}{c}{1.6} & \multicolumn{1}{c}{1.9} & \multicolumn{1}{c}{5.7} & \multicolumn{1}{c}{1.6}    & \multicolumn{1}{c}{\textbf{2.1}} & \multicolumn{1}{c}{4.6} \\
        \multicolumn{1}{l}{ULRSSM (A)} & \multicolumn{1}{c}{1.6}  & \multicolumn{1}{c}{2.2} & \multicolumn{1}{c}{\textbf{5.4}}  & \multicolumn{1}{c}{1.6} & \multicolumn{1}{c}{1.9} & \multicolumn{1}{c}{\textbf{5.1}} & \multicolumn{1}{c}{1.6}    & \multicolumn{1}{c}{2.2} & \multicolumn{1}{c}{\textbf{4.5}} \\
        \midrule
        \multicolumn{1}{l}{Hybrid ULRSSM (B)} & \multicolumn{1}{c}{1.4}  & \multicolumn{1}{c}{2.1} & \multicolumn{1}{c}{5.5}  & \multicolumn{1}{c}{1.4} & \multicolumn{1}{c}{1.8} & \multicolumn{1}{c}{5.4} & \multicolumn{1}{c}{1.5}    & \multicolumn{1}{c}{2.0} & \multicolumn{1}{c}{\textbf{3.6}} \\
        \multicolumn{1}{l}{Hybrid ULRSSM (A)} & \multicolumn{1}{c}{1.4}  & \multicolumn{1}{c}{\textbf{1.9}} & \multicolumn{1}{c}{\textbf{4.0}}  & \multicolumn{1}{c}{1.4} & \multicolumn{1}{c}{1.8} & \multicolumn{1}{c}{\textbf{5.1}} & \multicolumn{1}{c}{\textbf{1.4}}    & \multicolumn{1}{c}{\textbf{1.9}} & \multicolumn{1}{c}{3.7} \\
        \bottomrule
        \end{tabular} 

        }
\label{tab:near-iso}

\vspace{-2mm}
\end{table*}

\begin{table}[h!]
    \setlength{\tabcolsep}{3pt}
    \small
    \centering
    \caption{\textbf{Evaluation on partial-shape benchmark (SHREC’16).}
    Geodesic error ($\times 100$) on partial-to-partial matching tasks.}
    \label{tab:gradrot_partial}
    \begin{tabular}{@{}lcccc@{}}
        \toprule
        \multicolumn{1}{l}{Train}  & \multicolumn{2}{c}{\textbf{CUTS}} & \multicolumn{2}{c}{\textbf{HOLES}} \\ 
        \cmidrule(lr){2-3} \cmidrule(lr){4-5}
        \multicolumn{1}{l}{Test} & \textbf{CUTS} & \textbf{HOLES} & \textbf{CUTS} & \textbf{HOLES} \\ 
        \midrule
        ULRSSM (B) & 3.2 & 13.5 & \textbf{5.6} & \textbf{8.2} \\
        ULRSSM (A) & \textbf{2.6} & \textbf{13.0} & 5.8 & 8.7 \\
        \bottomrule
    \end{tabular}
\end{table}

\begin{table}[h!]
\centering
\footnotesize
\setlength{\tabcolsep}{4pt}
\renewcommand{\arraystretch}{1.1}
\caption{\textbf{Balanced Accuracy / Mean IoU (${\times 100}$) on partial-to-partial datasets.}  
Each entry reports \textit{balanced accuracy / mean IoU}, averaged across all test pairs.  }
\resizebox{\columnwidth}{!}{%
\begin{tabular}{@{}llccc@{}}
    \toprule
    \multicolumn{2}{c}{Methods} & \textbf{CP2P24} & \textbf{PSMAL} & \textbf{BeCoS} \\
    \midrule
    \multirow{2}{*}{EchoMatch (B)}& \textit{XYZ}      & 90.62 / 79.06 & 85.25 / 72.19 & 55.48 / 25.49 \\
        & \textit{DINOv2}   & 91.94 / 81.91 & 91.72 / 83.30 & \textbf{67.90} / 64.54\\ 
    \midrule
    \multirow{2}{*}{EchoMatch (A)}& \textit{XYZ}      & \textbf{91.03 / 80.10} & \textbf{85.99 / 72.71} & \textbf{60.13} / \textbf{52.40}\\
        & \textit{DINOv2}   & \textbf{93.53 / 84.72} & \textbf{92.75 / 84.75} & 67.05 / \textbf{64.68}\\
    \bottomrule
\end{tabular}
}
\label{tab:rotation_p2p}
\end{table}

\begin{figure}[h]%
    \centering
    \newcommand{\pckLineWidth}{1.5*1.5pt}
\newcommand{\plotWidth}{0.44*\linewidth*2.6}
\newcommand{\plotHeight}{0.37*\linewidth*2.3}
\newcommand{\pckTitle}{\textsc{\textbf{BeCoS}}}
\newcommand{\smCombAUC}{(39.79)}
\newcommand{\gcppsmAUC}{(30.72)}
\newcommand{\dpfmAUC}{(45.96)}
\newcommand{\oursAUC}{(59.18)}

\pgfplotsset{%
    label style = {font=\small},
    tick label style = {font=\small},
    title style =  {font=\normalsize},
    legend style={  fill= gray!10,
                    fill opacity=0.6, 
                    font=\small,
                    draw=gray!20, %
                    text opacity=1}
}
\begin{tikzpicture}[scale=0.7, transform shape]
	\begin{axis}[
		width=\plotWidth,
		height=\plotHeight,
		grid=major,
		title=\pckTitle,
		legend style={
			at={(0.01,0.98)},
			anchor=north west,
			legend columns=1},
		legend cell align={left},
	ylabel=\small{$\%$ Samples $<$Geo Dist Thresh},
        xmin=0,
        xmax=1,
        xlabel=\small{Geodesic Distance Threshold},
        ylabel near ticks,
        xtick={0, 0.25, 0.5, 0.75, 1, 1.25, 1.5},
        xticklabels={$0$, $0.25$, $0.5$, $0.75$, $1$, $1.25$,$1.5$},
        ymin=0,
        ymax=100,
        ytick={0, 20, 40, 60, 80, 100},
        yticklabels={$0$, $20$, $40$, $60$, $80$, $100$},
	]
\input{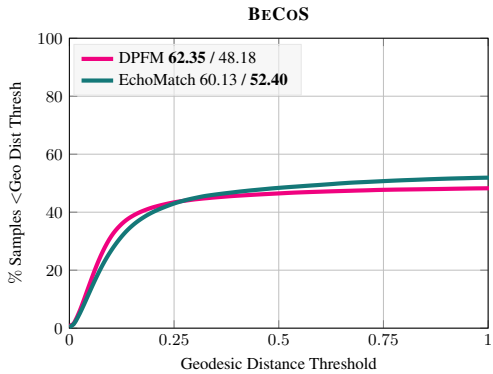}
\end{axis}
\end{tikzpicture}
\caption{\textbf{PCK curves} with Balanced Accuracy / Mean IoU ($\uparrow$) (legend values) for DPFM and EchoMatch on BeCoS with XYZ features. Despite lower IoU, DPFM achieves better geodesic accuracy in the low-error regime (0–0.25), a distinction reflected by balanced accuracy.
}
\label{fig:geo_errors}

\vspace{-4mm}
\end{figure}

\paragraph{Balanced Accuracy Metric.}
We evaluate balanced accuracy as a complementary metric for overlap prediction in partial-to-partial shape matching.
Table~\ref{tab:mIoU_balacc} reports IoU and balanced accuracy on CP2P24~\cite{attaiki2021dpfm,ehm2024partial}, PSMAL~\cite{ehm2024partial}, and BeCoS~\cite{becos}.
While EchoMatch achieves higher IoU across all settings, the geodesic error curves tell a more nuanced story: on BeCoS with \textit{XYZ} input, DPFM outperforms EchoMatch in the low-error regime of 0 to 0.25 geodesic distance despite its lower IoU, as shown in Fig.~\ref{fig:geo_errors}.
This distinction is reflected in the balanced accuracy (i.e., the low-threshold portion of the geodesic error curve), whereas the rightmost curve value corresponds directly to IoU.
Together, IoU, balanced accuracy, and the full geodesic error curve offer complementary views of the quality of overlap prediction.

\begin{table}[!t]
\centering
\footnotesize
\setlength{\tabcolsep}{4pt}
\renewcommand{\arraystretch}{1.1}
\caption{\textbf{Balanced Accuracy / Mean IoU (${\times 100}$) on partial-to-partial datasets.}  
Each entry reports \textit{balanced accuracy / mean IoU}, averaged across all test pairs.  
EchoMatch leads on IoU across all settings; however, on BeCoS with XYZ input, DPFM achieves higher balanced accuracy, reflecting its stronger performance in the low-error regime.
}
\begin{tabular}{@{}llccc@{}}
    \toprule
    \multicolumn{2}{c}{Methods} & \textbf{CP2P24} & \textbf{PSMAL} & \textbf{BeCoS} \\
    \midrule
    \multirow{2}{*}{DPFM}    
        & \textit{XYZ}      & 80.10 / 63.86 & 81.84 / 67.04 & \textbf{62.35} / 48.18 \\
        & \textit{DINOv2}   & 89.38 / 74.15 & 87.31 / 73.67 & 65.10 / 51.02 \\ 
    \midrule
    \multirow{2}{*}{EchoMatch} 
        & \textit{XYZ}      & \textbf{91.03 / 80.10} & \textbf{85.99 / 72.71} & 60.13 / \textbf{52.40} \\
        & \textit{DINOv2}   & \textbf{93.53 / 84.72} & \textbf{92.75 / 84.75} & \textbf{67.05 / 64.68} \\
    \bottomrule
\end{tabular}

\label{tab:mIoU_balacc}

\vspace{-4mm}
\end{table}

\section{Conclusion}
\label{sec:conclusion}

We have presented three contributions to improve the efficiency, understanding, and evaluation of deep functional map frameworks. 
Our batched functional map solver achieves up to a $33\times$ speedup over standard loop-based implementations while preserving exact solutions, directly benefiting any method built on this seminal formulation. 
Our analysis of two silently diverged DiffusionNet variants reveals complementary strengths across different deformation settings, motivating further study of learned tangent-plane transformations. 
Finally, adopting balanced accuracy for overlap prediction offers a simple yet informative complement to IoU across varying overlap ratios.
Looking forward, extending the batched solver formulation to non-orthogonal bases, such as those used in Hybrid Functional Maps~\cite{bastian2024hybrid}, remains an open direction for further acceleration. More broadly, a deeper investigation into the interaction between learned surface transformations and shape symmetries, non-isometries, and partiality could yield improvements beyond the implementation-level differences documented here. 
We release all implementations as part of \textit{DeepShapeMatchingKit}, an open-source codebase for deep shape matching, and hope it serves as a useful foundation for future research.

\YZ{
}

{\small
\bibliographystyle{ieeenat_fullname}
\bibliography{11_references}
}

\end{document}


\title{\paperTitle}
\author{\authorBlock}
\maketitlesupplementary

\appendix
\section{Appendix Section}
\label{sec:appendix_section}
Supplementary material goes here.

{\small
\bibliographystyle{ieeenat_fullname}
\bibliography{11_references}
}